\documentclass[10pt,journal,compsoc]{IEEEtran}
\newif\ifpeerreview

\peerreviewfalse

\usepackage[nocompress]{cite}
\usepackage{url}
\usepackage{amsmath,amssymb,graphicx}

\usepackage[switch]{lineno}

\usepackage{subcaption}
\usepackage{microtype}
\usepackage{soul}
\usepackage{enumitem}
\usepackage{comment} 
\usepackage{xcolor}
\usepackage{tabularx, booktabs}
\newcolumntype{H}{>{\setbox0=\hbox\bgroup}c<{\egroup}@{}}
\usepackage{array}
\usepackage{pifont} 
\newcommand{\revised}[1]{{#1}}
\usepackage[pagebackref,breaklinks,colorlinks]{hyperref}

\newcommand{\paperID}{0009}

\title{Time-of-Day Neural Style Transfer for Architectural Photographs}

\author{Yingshu Chen, 
        Tuan-Anh Vu, 
        Ka-Chun Shum, 
        Binh-Son Hua,
        and Sai-Kit Yeung
\IEEEcompsocitemizethanks{
\IEEEcompsocthanksitem Yingshu Chen, Tuan-Anh Vu, Ka-Chun Shum, and Sai-Kit Yeung are with the Hong Kong University of Science and Technology.\protect\\
\IEEEcompsocthanksitem Binh-Son Hua is with VinAI Research, Vietnam.}
}

\begin{document}

\IEEEtitleabstractindextext{%
\begin{abstract}
Architectural photography is a genre of photography that focuses on capturing a building or structure in the foreground with dramatic lighting in the background.
Inspired by recent successes in image-to-image translation methods, we aim to perform style transfer for architectural photographs. 
However, the special composition in architectural photography poses great challenges for style transfer in this type of photographs. Existing neural style transfer methods treat the architectural images as a single entity, which would generate mismatched chrominance and destroy geometric features of the original architecture, yielding unrealistic lighting, wrong color rendition, and visual artifacts such as ghosting, appearance distortion, or color mismatching. 
In this paper, we specialize a neural style transfer method for architectural photography. 
Our method addresses the composition of the foreground and background in an architectural photograph in a two-branch neural network that separately considers the style transfer of the foreground and the background, respectively. 
Our method comprises a segmentation module, a learning-based image-to-image translation module, 
and an image blending optimization module. 
We trained our image-to-image translation neural network with a new dataset of unconstrained outdoor architectural photographs captured at different magic times of a day, utilizing additional semantic information for better chrominance matching and geometry preservation. 
Our experiments show that our method can produce photorealistic lighting and color rendition on both the foreground and background, and outperforms general image-to-image translation and arbitrary style transfer baselines quantitatively and qualitatively. Our code and data are available at \url{https://github.com/hkust-vgd/architectural_style_transfer}.
\end{abstract}

}


\maketitle
\thispagestyle{empty} 

\IEEEraisesectionheading{
  \section{Introduction}\label{sec:introduction}
}
%
%
%
%
\IEEEPARstart{A}{rtificial} intelligence has been revolutionizing photography with high-fidelity image synthesis using generative modeling techniques, which has led to a wide range of new applications for image manipulation and editing. In this direction, style transfer is a special visual task that aims at generating aesthetically pleasant images in the style of a reference image. It has been well known that style transfer techniques can successfully extract artistic styles from famous paintings and seamlessly blend the styles into real photographs, generating novel images.

Architectural photography is a form of photography that captures subjects such as buildings into pictures with into visually pleasing pictures. In general, an architectural photograph often has a building in the foreground and a sky background captured at a specific time of a day that exhibits dramatic lighting. Taking architectural photographs has been so far a challenging task, requiring both skills and aesthetic senses of a professional photographer. 

The availability of neural networks has led to a new possibility: let a machine learn to generate realistic architectural photographs with new styles. In this paper, we realize time-of-day style transfer for architectural photographs using neural networks. Specifically, we consider styles of outdoor architecture photographs with dramatic lighting at magic times of a day. Magic times of a day refer to golden hours at sunset when the sun is falling close to the horizon, blue hours at twilight when the sun is below the horizon and nighttime after sunset and before sunrise without any sunlight. 
We call this problem the \emph{architectural style transfer} problem.

\begin{figure}[!t]
\centering
\includegraphics[width=1\linewidth]{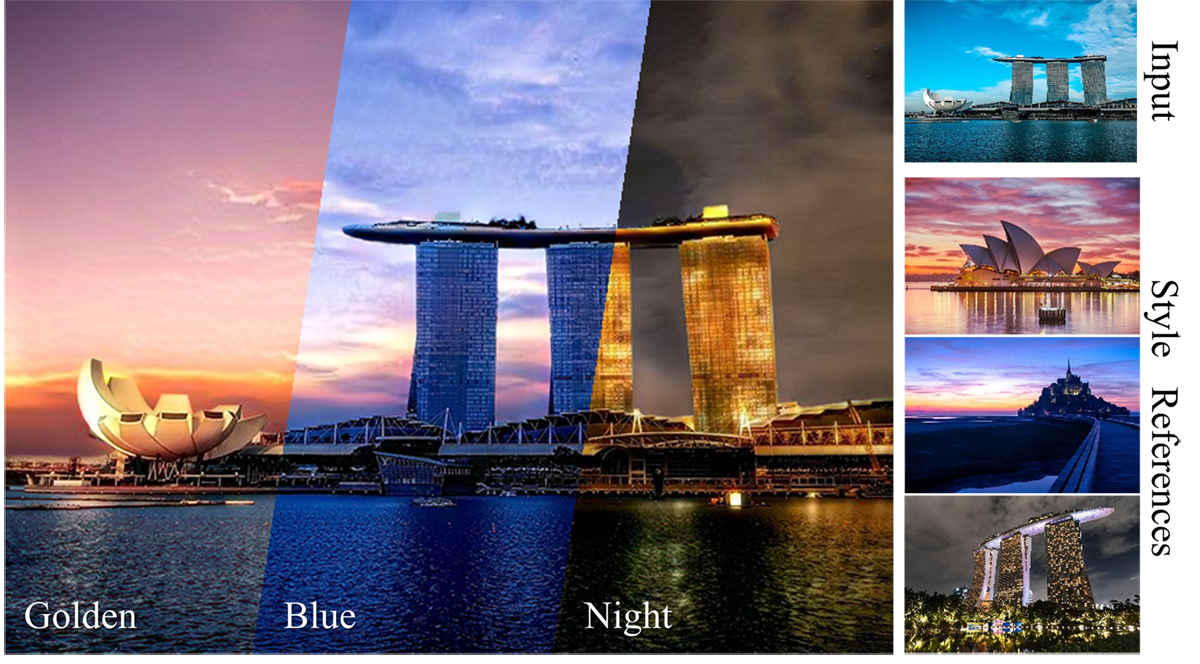}
\includegraphics[width=1\linewidth]{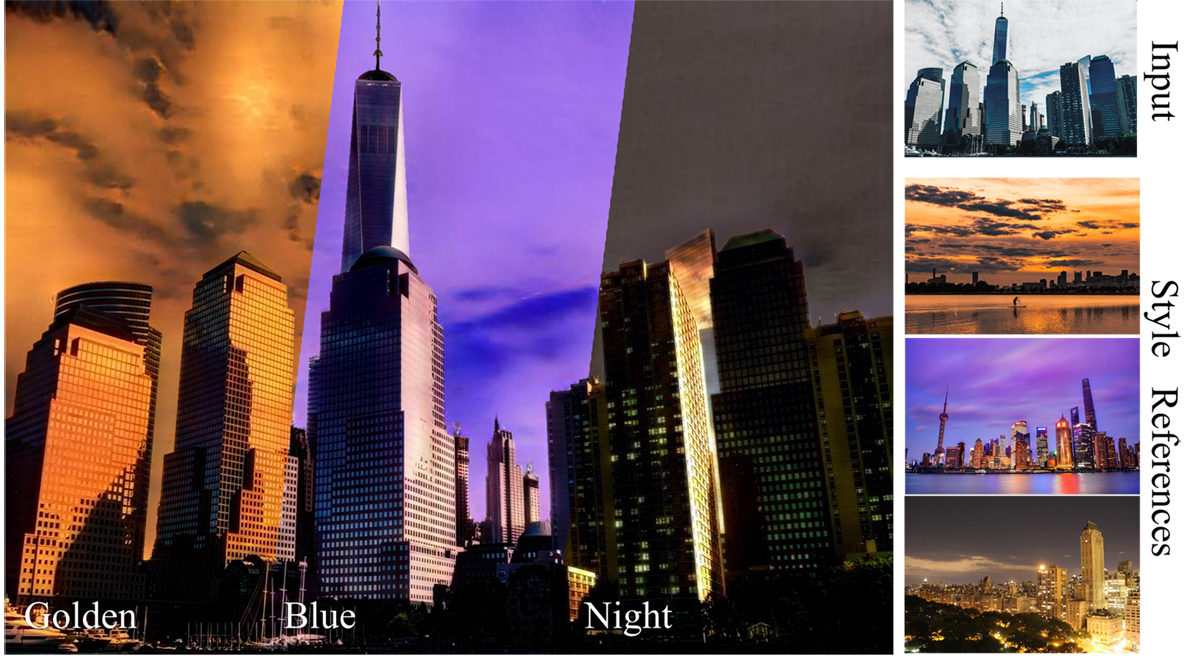}
\caption{Architectural style transfer at magic times of a day.}
\label{fig:teaser}
\end{figure}

Given the special composition of the foreground and background in architectural photographs, performing style transfer among these photographs is a challenging task.
In many cases, the style of the foreground and the background cannot be represented by a single latent space, e.g., the lighting on the building and the color and texture of the sky are very different. 
This causes existing image-to-image translation methods \cite{isola2017image, zhu2017unpaired,lee2018diverse,huang2018multimodal,liu2019few,chang2020domain} to fail because these methods treat the input image as a single entity; they are only good for global style transfer but fail to preserve image details or tend to exhibit visual artifacts when being used for architectural photographs. 
In the literature, there exist a few methods that consider style transfer for outdoor images, but they merely address natural landscapes \cite{karacan2019manipulating,anokhin2020high} or time-lapse videos \cite{nam2019end,cheng2020time}, which do not fit well to the composition in architectural photographs.

To overcome such challenges, we propose a new style transfer framework for architectural photography. 
We semantically disentangle the image content in an architectural photograph so that the style transfer can be done on the foreground (e.g., buildings) and the background (e.g., sky) respectively. 
For the foreground, we keep the geometry of static objects intact and transfer sufficient and appropriate style details. For the background, we transfer overall style including color and texture.
To realize this approach, we devise a two-branch neural network that handles the transfer for the foreground and background, respectively. 
We devise a set of loss functions for preserving the image details in the transfer.
To support training and testing, we also collect a new dataset of architectural photographs captured at different times of day.
Our method is validated with qualitative and quantitative comparisons with state-of-the-art image-to-image translation and arbitrary neural style transfer methods, which demonstrates the robustness of our method. 

In summary, our main contributions are:
\begin{itemize}[leftmargin=*]
  \item A new problem setting for style transfer: photorealistic style transfer for architectural photographs of different times of day; 
  \item An image-to-image translation framework with disentanglement representation that separately considers style transfer for image foreground and background respectively, accompanied with simple but effective geometry losses designed for image content preservation.
  \item A new dataset of architectural photographs and an extensive benchmark for architectural style transfer. 
\end{itemize}


\section{Related Work}

Our architectural style transfer problem is related to a large body of works in modern computer vision, namely style transfer, image-to-image translation, image relighting, and timelapse translation methods, which we discuss below. 

\subsection{Style Transfer}

Image style transfer has a long history in computer vision. Early techniques take user input as guidance, e.g., colorization methods~\cite{luan2007natural,ironi2005colorization} take user strokes to provide color mapping or segmentation cues. 
By using a pair of a reference (target) style image and an input (source) image, automatic methods aim at transferring the global style of the target image to the source image while preserving details of the source. 
This can be formulated as parametric algorithms such as color tone transfer~\cite{reinhard2001color}, data-driven search for style transfer mapping~\cite{shih2013data}, 
multi-level style feature transforms~\cite{li2018closed}, progressive stylization~\cite{yoo2019photorealistic}, image-optimization-based approaches \cite{gatys2016image, luan2017deep,penhouet2019automated,wang2021rethinking}. 
However, many of these methods usually require an iterative process or slow post-processing that is computationally expensive.

Prior works also explore style transfer using visual cues from semantic segmentation \cite{shih2014style,lu2017decoder,cheng2020time,liao2022semantic}, improving the photorealism based on the matting Laplacian \cite{levin2007closed,luan2017deep}, screened Poisson equation~\cite{mechrez2017photorealistic}, or photorealistic smoothing\cite{li2018closed} as a post-processing step. 
Our method belongs to the family of techniques in which segmentation information is explicitly used to address the style transfer problem.

With deep learning, neural networks can be trained and perform instant style or color transfer at inference time \cite{li2017universal, iizuka2016let, huang2017arbitrary,zhang2019deep,he2019progressive, park2019arbitrary, li2019learning, liu2021adaattn}.
Extended from the seminal work of arbitrary global style transfer with adaptive instance normalization (AdaIN)~\cite{huang2017arbitrary}, recent networks such as SANet~\cite{park2019arbitrary} and AdaAttN~\cite{liu2021adaattn} further introduce attention-aware mechanism to cover image local features and preserve better content appearance. Our network utilizes AdaIN~\cite{huang2017arbitrary} for style fusion; we empirically found AdaIN more robust for style transfer and attention-based networks like SANet or AdaAttN does not preserve photorealistic geometry.

\subsection{Image-to-image Translation}

Image-to-image translation is a task that aims to transfer an image from a source to a target domain, preserving its original content while having the characteristics of the target domain. Compared to traditional style transfer, this class of methods learns to perform the mapping from data from both domains. Image-to-image translation is, therefore, more general and can also be used for solving image colorization or style transfer problems. 
Typical methods of image-to-image translation include supervised methods such as pix2pix~\cite{isola2017image}, unsupervised methods such as CycleGAN~\cite{zhu2017unpaired} and UNIT~\cite{liu2017unsupervised}, multi-modal translation methods such as MUNIT~\cite{huang2018multimodal}, DIRT~\cite{lee2018diverse}, DSMAP~\cite{chang2020domain}, and more generic translators such as StarGAN v2~\cite{choi2020stargan} and  FUNIT~\cite{liu2019few} for handling multiple image classes. These methods are effective at translating style elements such as color, texture at a speedy inference time. 

An important factor to perform effective translation is to disentangle the content and style representation. In an encoder-decoder framework, recent methods~\cite{lee2018diverse,huang2018multimodal,liu2019few,chang2020domain,cheng2020time} inject both content code and style code generated from content encoder and style encoder respectively into the content generator to get the final transferred images with input content appearance but target style. 
Park \textit{et al.} \cite{park2020swapping} designed an autoencoder to swap textures between two images by disentangling structure and texture, which changes input content geometry and does not suit our task. 

In this work, we build our method upon the framework of image-to-image translation with disentanglement of content and style representation.
Moreover, the separation of luminance and chrominance of images in image-to-image translation is widely used in colorization~\cite{ho2020semantic} or color style transfer~\cite{martin2020nerf} tasks, and is beneficial to content integrity with luminance information intact. Therefore, we design losses that preserve high spatial frequency details of the image by considering the gradients of the luminance channel, so that the translation network can retain geometric details with high fidelity.

\subsection{Image Relighting and Timelapse Translation}

Our architectural style transfer problem is also relevant to some traditional image relighting methods. Particularly, Shih \textit{et al.} \cite{shih2013data} proposed an approach with a global to local mapping and local affine transformation for time-lapse appearance style transfer from video to photographs. It achieves an elegant color appearance change of an outdoor scene via rich appearance information from abundant time-lapse videos. However, this method requires a matched time-lapse video as a transfer reference and thus does not support arbitrary style transfer. More prior works such as Liu \textit{et al.}~\cite{liu2020learning}, Yu \textit{et al.}~\cite{yu2020self}, Laffont \textit{et al.}~\cite{laffont2012coherent} and Duch{\^e}ne \textit{et al.}~\cite{duchene2015multi} are tailored for image decomposition to achieve outdoor scene relighting. They require multi-view images~\cite{yu2020self,laffont2012coherent,duchene2015multi} or illumination-varying images~\cite{liu2020learning} to achieve intrinsic decomposition. These methods can relight scenes with general lighting conditions such as sunlight and shadow, but cannot transfer semantic color information. 
Later, Laffont \textit{et al.}~\cite{laffont2014transient} further explored more outdoor scene attribute transients and proposed a high-level example-based appearance transfer system.

In recent years, timelapse translation for images or videos can be done with deep learning on datasets from videos, landscapes, and street photos. Karacan \textit{et al.}~\cite{karacan2019manipulating} realize style transfer with specific attributes of natural scenes via generated style references produced by a deep scene generator. With different specific training data such as landscape photos~\cite{anokhin2020high}, photos in different time of sky~\cite{hold2019deep,xiong2018learning}, Google street views~\cite{anguelov2010google,zamir2014image}, timelapse transition can demonstrate in diverse ways, e.g., color and texture~\cite{anokhin2020high}, sunlight estimation~\cite{hold2019deep,xiong2018learning}, lighting and shadow conditions in street views~\cite{liu2020learning}. Animation video synthesis from a single outdoor photo is achieved by Endo \textit{et al.}~\cite{endo2019animating} predicting motion and appearance via convolutional neural network (CNN) models. Some latest works address time-lapse video synthesis, for examples, Nam \textit{et al.} \cite{nam2019end} and Cheng \textit{et al.}\cite{cheng2020time} trained end-to-end supervised models from time-lapse videos to synthesize time-lapse video from a single image.


\begin{figure*}[!t]
\centering
\includegraphics[width=1\linewidth]{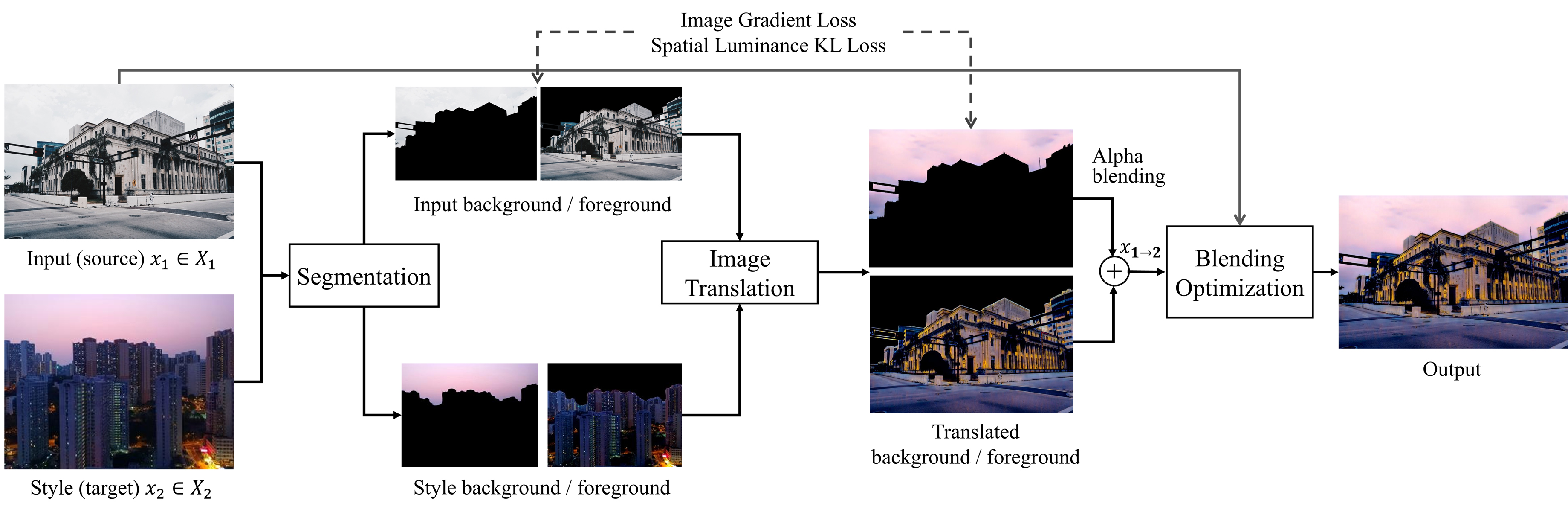}
\caption{Our architectural style transfer framework has three main modules: segmentation, image translation and blending optimization. Only style transfer in one direction ($X_1 \rightarrow X_2$) is illustrated. $x_1$ and $x_2$ are images from two domains ($X_1$ and $X_2$), $x_1$ as high-fidelity source input, $x_2$ as target style reference. After segmentation, foreground and background images are fed into the translation network respectively. The translated image $x_{1\rightarrow2}$ (or $c_{style}$) and the input source $x_1$ can be further post-processed by the blending optimization module. More details are in Section~\ref{method_network},~\ref{method_geometry}, and~\ref{method_blend}.}
\label{fig:framework}
\end{figure*}

\section{Background and Motivation}
The primary emphasis of an architectural photograph is the harmonious rendition of a building structure in the foreground on an aesthetically pleasing background. Given an input architectural image (source) and a style architectural image (target), we aim to transfer the characteristics of the target to the source while preserving the content and structure of the source image. The transferred characteristics include the color and texture rendition in both the foreground and the background of the target. An example of the source, target, and the transferred result can be seen in Figure~\ref{fig:teaser}. We call this problem the architectural style transfer problem. 

The foreground and background in architectural photographs pose unique challenges in this domain-specific style transfer problem. At first, we are tempted to directly use recent methods for generic neural style transfer and image-to-image translation to solve this problem, but we soon realized that these methods barely work up to our expectations, and there remain the following challenges. Existing image-to-image translation methods treat the image as a single entity without knowing the foreground and background. The lack of this inductive bias in architectural photography makes these methods perform not as effectively. We found that while these methods are efficient in transferring the global style, they tend to operate poorly in preserving geometry details of the foreground, having visual artifacts, or producing unfaithful results with mismatching color (e.g., Fig.~\ref{fig:exp_seg} (c)). It is, therefore, necessary to derive a specific method for architectural style transfer that takes all such issues into account. 

\section{Proposed Method}

\subsection{System Overview}

We design a specific method for image-to-image translations of architectural photographs. We take the inductive bias from architectural photography into account by having the neural networks learn to transfer the foreground and background styles, respectively. This learning bias, while seemingly trivial, provides a strong constraint for architectural style transfer as it allows us to define the semantic correspondences between the input and the style image, effectively model the characteristics of architectural photographs because the style of the foreground and background can be vastly different and diverse.

Our framework has three main modules: image segmentation, image translation, and image blending. Given an input image and a style image, we first segment their background and foreground for data preprocessing, and then train the image translations separately for the foreground and background. 
To train the neural networks, we propose a new geometry loss to preserve structural details that are vital in architectural images. Particularly, we aim to preserve the geometry contour (\textit{Image Gradient loss}) and spatial luminance distribution (\textit{Spatial Luminance KL-divergence loss}) for image foreground translation. 
With the illumination density constraint, empirically, the appearance information is well retained. 

To produce the final result, we blend the predicted foreground and background using the original high-fidelity input source as a geometric constraint. We intentionally let this step be fixed and not trainable as we found that such a post-processing step can already provide satisfactory results. 
An overview of our framework is presented in Figure \ref{fig:framework}.


\subsection{Semantic Correspondences} \label{method_semantic} 

We determine the foreground and background of the architectural images by a semantic segmentation model. Particularly, we segment the input and the style images into the background (i.e., sky) and foreground (other elements such as buildings, trees, rivers, etc.). This allows us to build high-level semantic correspondences between the input and the style image to perform the transfer on each correspondence, respectively. 
Our semantic segmentation is built upon an encoder-decoder model as follows. We use ResNet-50-dilated \cite{HeZRS15} for the encoder, and the pyramid pooling module with loss optimization from PSPNet \cite{zhao2017pyramid} for the decoder. 
We use the official pretrained model trained on the ADE20K dataset \cite{zhou2018semantic}.

Given the pretrained semantic segmentation model, we preprocess our training images by applying the model to predict the masks that indicate the foreground and background. We store all the masks and use them to separate the foreground and background for our training. At inference, we have the flexibility to allow users to use the pretrained segmentation model to automatically segment the inputs or manually provide their masks. 
We empirically observed that our training tolerates segmentation imperfection at a certain extent. We discuss some failure case due to imperfect segmentation in Sec.~\ref{sec:limitation} and Fig.~\ref{fig:fail_case}.

\begin{figure}[!t]
\centering
\includegraphics[width=0.48\textwidth]{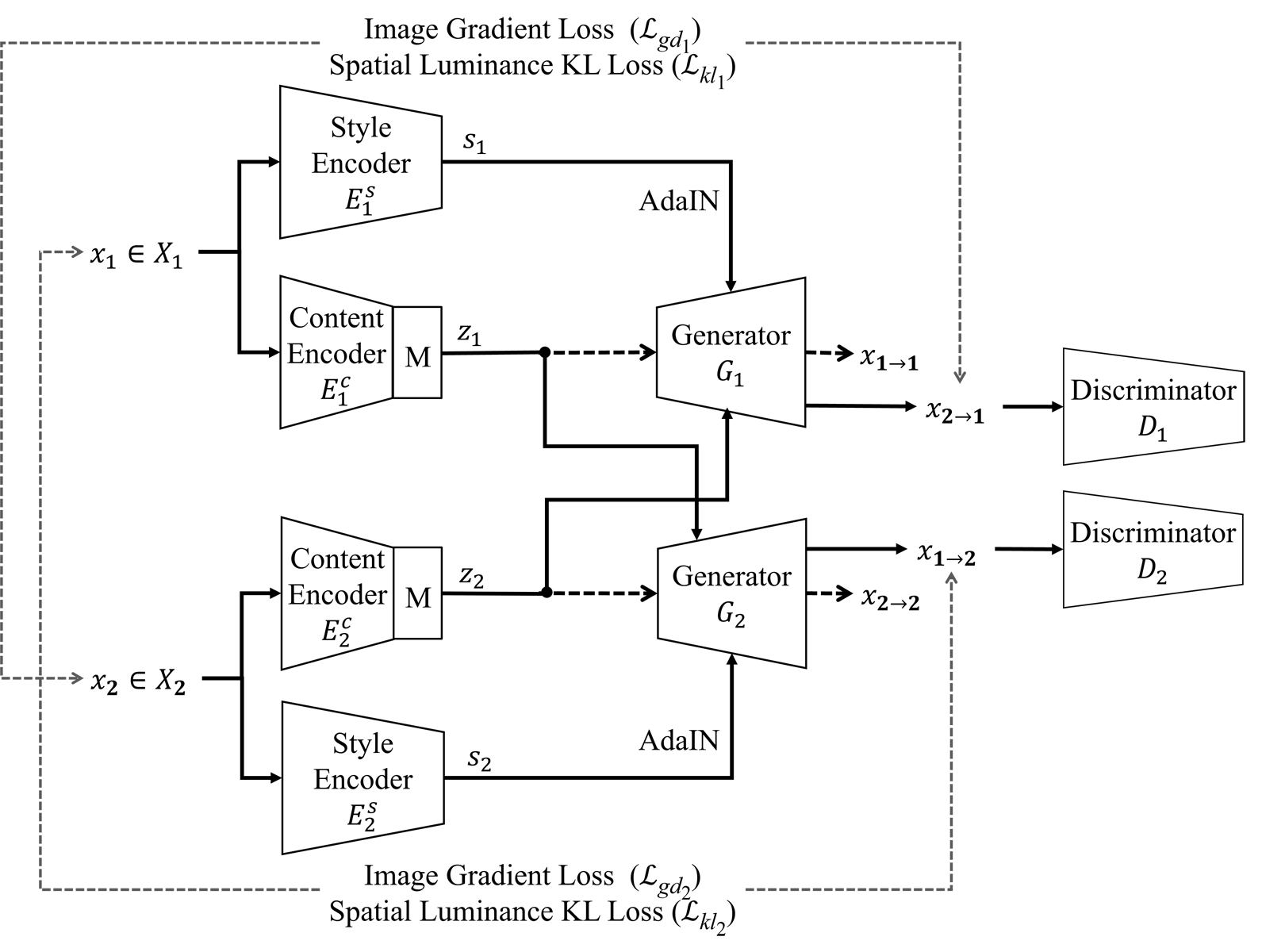}
\caption{The common image-to-image translation network architecture. $x_1$ and $x_2$ are images from two domains ($X_1, X_2$). $x_{1\rightarrow2}$ and $x_{2\rightarrow1}$ are style transferred outputs between two domains. $x_{1\rightarrow1}$ and $x_{2\rightarrow2}$ are reconstruction outputs. We use the same architecture for foreground and background translation.}
\label{fig:translation_framework}
\end{figure}

\subsection{Neural Network Architecture} \label{method_network}
Given the source domain $X_1$, the target domain $X_2$, and a pair of image $x_1 \in X_1, x_2 \in X_2$, our goal is to develop an image-to-image translation network to transform $x_1$ to the target domain, i.e., the translated image should resemble in style of $x_2$ while preserving the details in $x_1$. 
Our method is based on the disentanglement of style and content under the same separation domains assumption as~\cite{lee2018diverse,huang2018multimodal,chang2020domain}. 
It assumes that each image belongs to a shared domain-invariant content latent space $\mathcal{D}^c$ but a different domain-specific style latent space $\mathcal{D}^s$. 
An overview of our method is shown in Fig.~\ref{fig:translation_framework}.

Particularly, to transfer style from source domain $X_1$ to target domain $X_2$, we employ two encoders $E_1^c$, $E_2^s$ to embed content and style features respectively into latent spaces $\mathcal{D}^c$ and $\mathcal{D}^s$, and a decoder $G_2$ to generate results with specified content and style. 
We further employ a mapping module $M$ followed the design of DSMAP~\cite{chang2020domain} to map the content latent space $\mathcal{D}^c$, which is domain invariant, to become domain-specific for better generation in the target domain. 

Let $c_1 = E_1^c(x_1) \in \mathcal{D}^c$ be the domain-invariant content code of image $x_1$, and $z_1 = M(c_1)$  be the domain-specific content code after mapping.
The style code can be extracted as $s_2=E_2^s(x_2) \in \mathcal{D}^s$.
In the generator $G_2$, we apply adaptive instance normalization (i.e., AdaIN \cite{huang2017arbitrary}) for style transfer. 
Finally, the latent content and style codes are fed into the generator to obtain a new image $x_{1\rightarrow2} = G_2(z_1, s_2) \in X_2$ that has the content of $X_1$ and the style of $X_2$. 
Likewise, the translated image $x_{2\rightarrow1}$ from domain $X_2$ to $X_1$ can be obtained via $x_{2\rightarrow1} = G_1(z_2, s_1)$.

Accordingly, we have two discriminators $D_1$ and $D_2$ to discriminate the real images and generated images in each domain $X_1$ and $X_2$, respectively. Similar to Huang \textit{et al.} \cite{huang2018multimodal}, we employ the multi-scale discriminator architecture.
The details of our network architecture are illustrated in the supplementary.

We separately apply the same network architecture to perform style transfer for both the foreground and the background, respectively. We empirically found that such a separation is necessary and robust because the styles of the foreground and background are vastly different. Both network branches for the foreground and the background share the same set of training objectives. We leave the investigation of joint training both branches as future work.

\subsection{Training Objectives} \label{method_geometry} 
We design a set of training objectives that can work for both the foreground and the background.
We train both translation networks using unpaired data with the reconstruction loss, cycle-consistency loss, and adversarial loss. 
Particularly, to preserve high-frequency geometry information of foreground, we presume the luminance of an image contains both geometry and illumination information, and devise the geometry losses (i.e., \textit{Image Gradient loss and Spatial Luminance KL-divergence loss}) to guide the generator to reproduce high-frequency content of the source.  

Here we detail the losses by assuming the transfer direction to be from domain $X_1$ to domain $X_2$.

\noindent\textbf{Image Gradient Loss}. 
Image gradient can well represent edges of objects in an image. Preserving the good gradient attribute of an image to some extent guarantees the photorealism and fidelity~\cite{mechrez2017photorealistic}. Our image gradient loss for $x_{1\rightarrow2}$ is:
\begin{equation}
\label{eq:lgd1}
\mathcal{L}_{gd_1} =\mathbb{E}_{x_1,x_2} \left[ \left\| \nabla( Y(x_{1\rightarrow2}))-\nabla(Y(x_1)) \right\|_1 \right],
\end{equation}
where $\mathbb{E}[.]$ is the expectation operator, $\nabla(\cdot)$ is the image gradient, $Y(x)$ is luminance of image $x$. Following ITU-R BT.601 conversion standard \cite{wikipedia_2022} to get Y channel values from RGB channels, we define the luminance by
\begin{equation}
    Y = 0.299 \times R + 0.587 \times G + 0.114 \times B.
\end{equation}
Here R, G and B are image values in RGB channels.

\noindent\textbf{Spatial Luminance KL Divergence Loss}. 
The relative entropy or the so-called Kullback-Leibler divergence (KL divergence), is a useful distance measure for continuous distributions. To constrain the \revised{geometry} luminance distribution to the \revised{source input image},
we apply KL divergence loss on the luminance channel of output $x_{1\rightarrow2}$ and \revised{input $x_1$}:
\begin{equation}
\label{eq:lkl1}
\mathcal{L}_{kl_1} = \mathbb{E}_{x_1,x_2}\left[\mathrm{KL}(Y(x_{1\rightarrow2})\parallel Y(\revised{x_1}))\right]
\end{equation}
where each value of luminance $Y(x)$ is normalized to $[0, 1]$ when calculating the loss; $\mathrm{KL}(p\parallel q)=\sum_x p(x)\log[ p(x)/q(x) ]$ measures the KL divergence between a distribution $p$ and a reference distribution $q$. This loss constrains the model to generate images in illumination distribution of target domain (${X_2}$).

We call the total of image gradient loss and the KL divergence loss the \emph{geometry loss} as it can improve the geometry quality while transferring the illumination faithfully. 

\noindent\textbf{Reconstruction Loss}.
We utilize the same concept of bidirectional reconstruction loss in \cite{huang2018multimodal} for image reconstruction loss, which involves image self-reconstruction loss $\mathcal{L}_x$, content latent code  reconstruction loss $\mathcal{L}_c$ and style latent code reconstruction loss $\mathcal{L}_s$. And we have reconstruction loss of domain-specific content latent code $\mathcal{L}_{z}$ same as that in \cite{chang2020domain}:
\begin{align}
\mathcal{L}_{x_1}&=\mathbb{E}_{x_1}[\parallel x_{1\rightarrow1} - x_1 \parallel_1],\\
\mathcal{L}_{c_1}&=\mathbb{E}_{x_1,x_2}[\parallel E_1^c(x_{1\rightarrow2}) - E_1^c(x_1) \parallel_1],\\
\mathcal{L}_{s_2}&=\mathbb{E}_{z_1,r}[\parallel E_2^s(G_2(z_1, r))) - r \parallel_1],\\
\mathcal{L}_{z_1}&=\mathbb{E}_{x_1,x_2}[\parallel M(E_1^c(x_{1\rightarrow2})) - M(E_1^c(x_1))) \parallel_1]
\end{align} 
where $x_{1\rightarrow1}=G_1(z_1, s_1)$, $G_2(z_1, r)$ is generation with appearance of $x_1$ and random style $r$ in style space of $X_2$, $r$ is a random value drawn from a Gaussian distribution \cite{lee2018diverse} to ensure diversity of the style embedded codes and multi-modal translations.

On one side, we consider the reconstruction of the input source image; on the other side, we take into account the reconstructions of domain-specific content latent code, style latent code, and domain-invariant content latent code, respectively from the content encoder $E^c_1$ and style encoder $E^s_2$ and content space mapping $M$. 

\noindent\textbf{Cycle Consistent Loss.}
For unsupervised GANs, Zhu \textit{et al.}~\cite{zhu2017unpaired} proposed the cycle-consistent loss, which is widely accepted in unsupervised image translation tasks. We compute the cross-cycle consistency loss $\mathcal{L}_{cycle}$ between input $x_1$ and the reconstruction from domain $X_1$ to domain $X_2$ and back to $X_1$ again (i.e., $X_{1\rightarrow2\rightarrow1}$):
\begin{align}
\mathcal{L}_{cycle_1}&=\mathbb{E}_{x_1,x_2}\left[\parallel G_{1 \rightarrow 2 \rightarrow 1}(x_1, x_2), - x_1 \parallel_1\right].
\end{align}

\noindent\textbf{Adversarial Loss.}
We also adopt the adversarial loss $L_{adv}$ of LSGAN~\cite{mao2016least} between discriminators and generators:
\begin{align}
\mathcal{L}^{D_2}_{adv_1}&=\mathbb{E}_{x_1,x_2}\left[ \frac{1}{2}(D_2(x_1)-1)^2 + \frac{1}{2}D_2(G_2(z_1,s_x))^2\right],\\
\mathcal{L}^{G_2}_{adv_1}&=\mathbb{E}_{x_1,x_2}\left[ \frac{1}{2}(D_2(G_2(z_1,s_x)-1)^2\right],
\end{align}
where $D_2$ is the discriminator for images in target domain $X_2$, $s_x = \{s_2, r\}$.

\noindent\textbf{Total Loss.}
All encoders, generators and discriminators are trained simultaneously with bidirections. We get the final total loss for the generator as
\begin{align}
\mathcal{L}_{total} &= \lambda_{x} \mathcal{L}_x + \lambda_{c} \mathcal{L}_c
+\lambda_{s} \mathcal{L}_s + \lambda_{z} \mathcal{L}_{z} \\ \nonumber
&+\lambda_{cycle} \mathcal{L}_{cycle} +\lambda_{adv} \mathcal{L}_{adv} +\lambda_{gd} \mathcal{L}_{gd} +\lambda_{kl} \mathcal{L}_{kl},
\end{align}
where $\lambda$'s are hyperparameters to balance the losses. Each loss contains losses in both directions
\begin{align}
\mathcal{L}_\ast=\mathcal{L}_{\ast_1}+\mathcal{L}_{\ast_2}
\end{align} 
where $\ast \in \{x, c, s, cs, cycle, adv, gd, kl\}$. For training with background we set $\lambda_{gd}=\lambda_{kl}=0$. 

\begin{figure}[!t]
\centering
\includegraphics[width=1\linewidth]{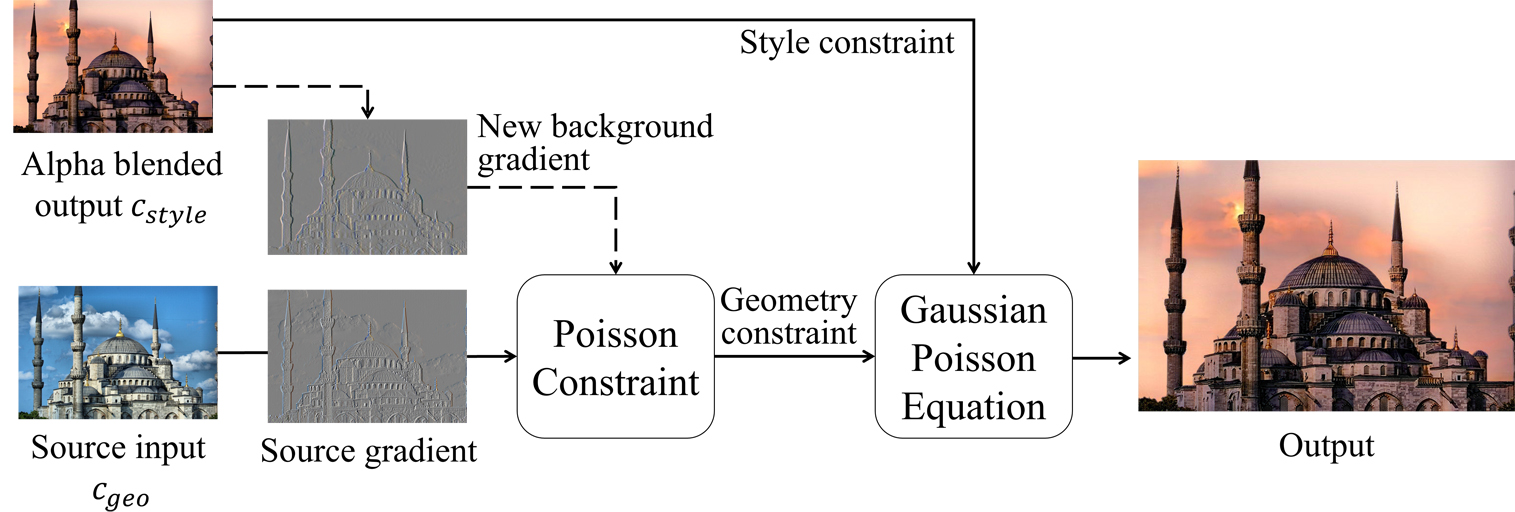}
\caption{Blending optimization. Alpha blended image $c_{style}$ (e.g., $x_{1\rightarrow2}$) works as style constraint, or meanwhile as background geometry constraint. The high-fidelity source input $c_{geo}$ (e.g., $x_1$) constrains the precise image contour, enhancing overall photorealism.}
\label{fig:blend_framework}
\end{figure}

\noindent\textbf{Hyperparameter Settings}
In training, we adapt the Adam Optimizer with an initial learning rate of $1\times10^{-4}$, $\beta_1=0.5$, $\beta_2=0.999$.
For foreground and background training, we set different hyperparameters for geometry loss, remaining weights the same for all our models. Empirically, for foreground training, we set the weight of image gradient loss $\lambda_{gd}=5$, weight of spatial luminance KL divergence loss $\lambda_{kl}=5$, while for background training, we set both of them zero to let the generator learn to change the background texture. For other losses, we empirically set $\lambda_{x}=10$, $\lambda_{c}= 2$, $\lambda_{cs}= 2$, $\lambda_{s}= 10$, $\lambda_{cc}=5$, $\lambda_{adv}=1$. 
We set the batch size to 2 for training. Each model for foreground and background was trained for 200k iterations, respectively. 

\begin{table*}[!t]
\centering
\renewcommand*{\arraystretch}{1.1}
\setlength{\tabcolsep}{2.8pt}
\begin{tabular}{l | c c c c c | c c c c H | c }
\toprule
 & DRIT++\cite{lee2018diverse,lee2020drit++} & MUNIT\cite{huang2018multimodal} & FUNIT\cite{liu2019few}  & DSMAP\cite{chang2020domain} & StarGANv2\cite{choi2020stargan} & AdaIN\cite{huang2017arbitrary} & SANet\cite{park2019arbitrary} & AdaAttN\cite{liu2021adaattn} & LST\cite{li2019learning} & WCT$^2$ & Ours \\
\midrule
e-SSIM$\uparrow$ & 0.5214 & \underline{0.5653} & 0.4959 & 0.4790 & 0.4778 & 0.4962 & 0.4854 & 0.5194 & 0.4903 & WCT2 & \textbf{0.6359}\\
Acc$\uparrow$  & 0.8903 & 0.8678 & 0.77.14 & \underline{0.9106} & 0.8788 & 0.7352 & 0.6193 & 0.6443 & 0.7071 & WCT2 & \textbf{0.9486}\\
IS$\uparrow$ & 2.6160 & 2.5916 & 2.5903 & \underline{2.6580} & 2.6088 & 2.4082 & 2.1062 & 2.0928 & 1.7299 & WCT2 & \textbf{2.7290} \\
IoU$\uparrow$ & 0.6915 & \textbf{0.7382} & 0.5473 & 0.4975 & 0.4100 & 0.6642 & 0.7183 & 0.6532 & 0.6264 & 0.7673 & \underline{0.7257} \\
\bottomrule
\end{tabular}
\caption{Evaluation results of Daytime to Golden, Blue and Nighttime Hour translations. \textbf{Bold} and \underline{underlined} text indicates the best and 2nd best result, respectively.}
\label{table:exp_metrics}
\end{table*}
\begin{table*}[!t]
\centering
\renewcommand*{\arraystretch}{1.1}
\setlength{\tabcolsep}{2.8pt}
\begin{tabular}{l | c c c c c | c c c c H | c }
\toprule
 & DRIT++\cite{lee2018diverse,lee2020drit++} & MUNIT\cite{huang2018multimodal} & FUNIT\cite{liu2019few}  & DSMAP\cite{chang2020domain} & StarGANv2\cite{choi2020stargan} & AdaIN\cite{huang2017arbitrary} & SANet\cite{park2019arbitrary} & AdaAttN\cite{liu2021adaattn} & LST\cite{li2019learning} & WCT$^2$ & Ours-opt\\
\midrule
e-SSIM$\uparrow$ & 0.4939 & \underline{0.5074} & 0.5003 & 0.4911 & 0.4935 & 0.4863 & 0.4921 & 0.4706 & 0.4815 & WCT$^2$ & \textbf{0.8094}\\
Acc$\uparrow$ & 0.8586 & 0.8507 & 0.7532 & \underline{0.8721} & 0.8391 & 0.7019 & 0.5332 & 0.5973 & 0.6623 & WCT$^2$ & \textbf{0.9007} \\
IS$\uparrow$ & 2.5528 & 2.5236 & 2.5227 & \underline{2.5572} & 2.5087 & 2.2646 & 1.8310 & 1.9978 & 2.1587 & WCT$^2$ & \textbf{2.6127}  \\
IoU$\uparrow$ & 0.6997 & 0.7228 & 0.7143 & 0.6972 & 0.6902 & 0.7083 & \underline{0.7369} & 0.7023 & 0.7053 & WCT$^2$ & \textbf{0.7715}\\
\bottomrule
\end{tabular}
\caption{Evaluation results of Daytime to Golden, Blue and Nighttime Hour translations with blending optimization applied to all methods. \textbf{Bold} and \underline{underlined} text indicates the best and 2nd best result, respectively.}
\label{table:exp_metrics_opt}
\end{table*}

\subsection{Image Blending Optimization} \label{method_blend}
After image translation, we get two generated images (foreground and background). Foreground and background generated images are integrated again with the segmentation mask from the segmentation module using alpha blending.

We apply a similar strategy to \cite{wu2019gp} for blending optimization (see Fig.\ref{fig:blend_framework}), which helps restore the original gradient. Instead of training a new GAN to generate a relatively low-resolution color constraint image as described in~\cite{wu2019gp}, we apply our translated image as the style constraint ($c_{style}=x_{1\rightarrow2}$). With realistic low-resolution style constraint ($c_{style}$) and the high-fidelity source image ($c_{geo}=x_1$) with perfect geometry, we iteratively optimize the Gaussian Poisson Equation~\cite{wu2019gp} and finally retrieve source geometry while the transferred style is preserved. Differently, to retain novel generated background textures (e.g., new cloud texture), the style constraint image is used to extract a new background gradient for blending optimization.
Empirically, 1 or 2 iterations are enough for high fidelity restoration.

\section{Experiments}

In this section, we first introduce the dataset used for training and evaluation, and then the baselines and evaluation metrics. Next, quantitative and qualitative comparisons are reported and discussed. In addition, the ablation study results are illustrated to validate the effectiveness of our framework design. Finally, we show the comparisons of the proposed method (deep learning base) with traditional methods (non-deep learning base).

\subsection{Time-lapse Architectural Dataset}

To better achieve style transition of different times in the day for architectural images, we collected 21,291 high-resolution exterior architectural photos for training. The training photos include 16,908 unpaired landmark photos in the wild and 4,383 extracted frames from 110 time-lapse videos of outdoor scenes from~\cite{shih2013data}. The evaluation set consists of 1,003 photos of high fidelity collected from public domains~\cite{pexels, pikwizard, unsplash}. We manually filtered out low-resolution or unattractive images, labeled images into four classes (day, golden, blue, night), and pre-processed images with proper cropping and resizing before training. All photographs of the sky and main architecture(s) have different geometry, luminance, and chrominance changes. The sky has a dynamic appearance as time elapses, while the architectures keep static geometry. Therefore, an outstanding time-of-day architectural style transfer should accomplish matched style transfer semantically. 

More details can be checked in the supplementary. We distribute our training dataset upon request for research purposes and release our evaluation set publicly.

\subsection{Experiment Setup}
 
\noindent\textbf{Baselines.} We qualitatively and quantitatively compare our results to state-of-the-art image-to-image translation baselines, i.e., DRIT++\cite{lee2018diverse,lee2020drit++}, MUNIT\cite{huang2018multimodal}, FUNIT\cite{liu2019few} and DSMAP\cite{chang2020domain}, and StarGANv2\cite{choi2020stargan}; neural style transfer methods including AdaIN~\cite{huang2017arbitrary}, LST~\cite{li2019learning}, SANet~\cite{park2019arbitrary}, AdaAttN~\cite{liu2021adaattn}. For ablation study, our method also compare to our variants without proposed geometry loss, and our model trained without segmentation.

\noindent\textbf{Implementation Details.} 
We trained all models until the total generation loss did not decrease noticeably. All baselines are trained using whole images, while our work uses segmented images with different hyperparameter settings for foreground and background training. 

At the training time of baselines and ours, all images are resized into 286$\times$286 and randomly cropped to 256$\times$256 with random horizontal flipping. At inference time, photos are resized with the shortest side to 256 pixels or 512 pixels to generate transferred results. All models are trained in a resolution of 256$\times$256. Our networks are fully convolutional and support input images of arbitrary resolutions. For quantitative evaluation, results are evaluated in 256$\times$ at smaller side. For qualitative comparison, we display results of 512$\times$ at smaller side. The teaser Fig~\ref{fig:teaser} show results of 1024$\times$ at smaller side.

All methods use a single NVIDIA GeForce RTX 2080 Ti GPU for training except that FUNIT utilizes 4 GPUs. 
FUNIT, StarGANv2 and all neural style transfer approaches trained with daytime images as source and all four classes together as targets. We use the default training configuration for all baselines except that we increase the weight of content loss in AdaIN (content : style = 3 : 10) and SANet
(content : style =  2 : 1) 
and the weight of local feature loss in AdaAttN (local : global = 1 : 2) to enhance the geometry of the stylized images. 

The unseen evaluation set apart from the training set is used for result inference. The main exploratory experiments are three types of style transfers, i.e., daytime to golden hours, daytime to blue hours, and daytime to nighttime. In the evaluation set, all daytime images get transferred with every style reference image in each target-style set (i.e., golden, blue, and night), and we got totally over 200k results from each method for quantitative evaluation.

\noindent\textbf{Performance Metrics.} 
The quantitative evaluation involves generation accuracy and diversity, geometry and appearance preservation, and semantic style transfer.

We trained an InceptionV3\cite{szegedy2016rethinking} classifier using our dataset with three target domain labels (i.e., golden, blue, and night). We evaluate the generation top-1 accuracy and Inception Score (IS) \cite{salimans2016improved} which indicate how realistic and diverse the generation is.

To quantitatively evaluate geometry and appearance similarity, we utilize Structural Similarity Index (SSIM)\cite{wang2004image}.
Similar to \cite{yoo2019photorealistic}, we calculate \textit{Edge Conditioned SSIM} (edge-SSIM), which computes image structural similarity (SSIM) between Canny edge detected maps of images \cite{canny1986computational}, alleviating luminance influence on geometry. 

Intersection over Union (IoU) between input and output is used to evaluate structure preservation and visual recognizability in foreground and background. We use the same segmentation model used in our framework.

\begin{figure}[!t]
\centering
\small
\includegraphics[width=0.24\linewidth,height=0.18\linewidth]{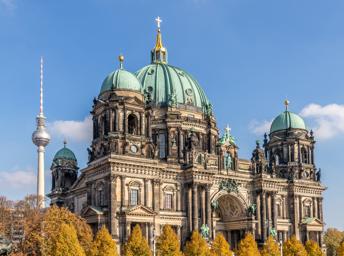}
\includegraphics[width=0.24\linewidth,height=0.18\linewidth]{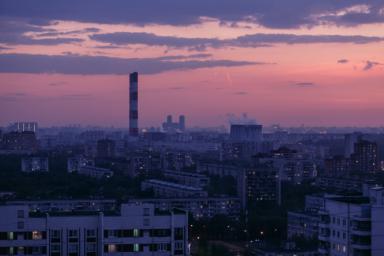}
\includegraphics[width=0.24\linewidth,height=0.18\linewidth]{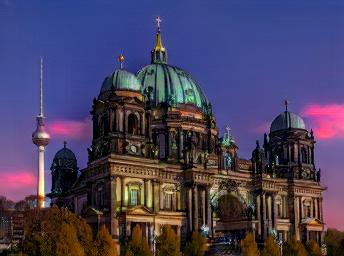}
\includegraphics[width=0.24\linewidth,height=0.18\linewidth]{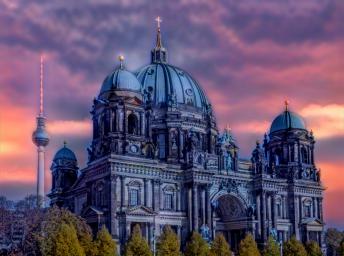}

\begin{subfigure}[b]{0.24\linewidth}
    \centering
    \caption{Input}
\end{subfigure}
\begin{subfigure}[b]{0.24\linewidth}
    \centering
    \caption{Style}
\end{subfigure}
\begin{subfigure}[b]{0.24\linewidth}
    \centering
    \caption{Whole}
\end{subfigure}
\begin{subfigure}[b]{0.24\linewidth}
    \centering
    \caption{Segmented}
\end{subfigure}

\caption{Qualitative comparison on segmentation condition. (a) Input image. (b) Style reference. (c) Style transfer result from our network trained on whole images. (d) Result from our network trained on segmented images.}
\label{fig:exp_seg}
\end{figure} 

\begin{figure}[!t]
\centering
\small
\def\wsc{0.32}
\def\hsc{0.22}
\includegraphics[width=\wsc\linewidth,height=\hsc\linewidth]{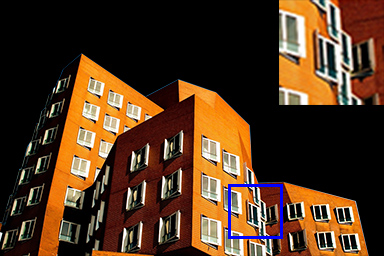}
\includegraphics[width=\wsc\linewidth,height=\hsc\linewidth]{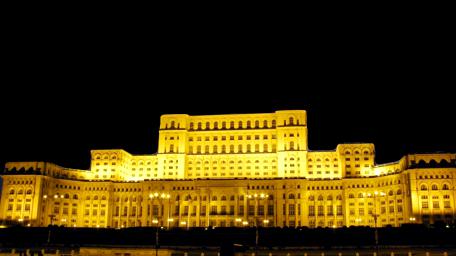}
\includegraphics[width=\wsc\linewidth,height=\hsc\linewidth]{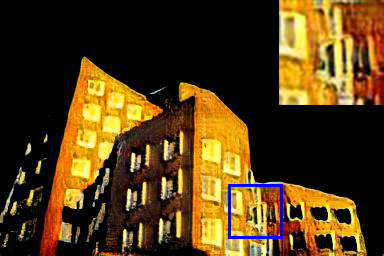}
\begin{subfigure}[b]{\wsc\linewidth}
    \caption{Input}
\end{subfigure}
\begin{subfigure}[b]{\wsc\linewidth}
    \centering
    \caption{Style reference}
\end{subfigure}
\begin{subfigure}[b]{\wsc\linewidth}
    \caption{w/o $\mathcal{L}_{gd}+\mathcal{L}_{kl}$}
\end{subfigure}

\vspace{0.5em} 
\includegraphics[width=\wsc\linewidth,height=\hsc\linewidth]{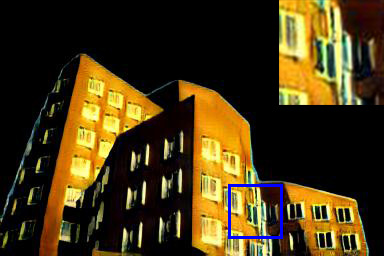}
\includegraphics[width=\wsc\linewidth,height=\hsc\linewidth]{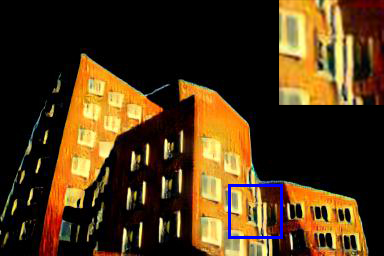}
\includegraphics[width=\wsc\linewidth,height=\hsc\linewidth]{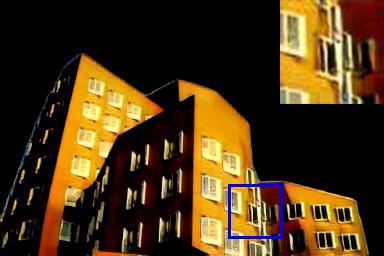}
\begin{subfigure}[b]{\wsc\linewidth}
    \centering
    \caption{w/o $\mathcal{L}_{kl}$}
\end{subfigure}
\begin{subfigure}[b]{\wsc\linewidth}
    \centering
    \caption{w/o $\mathcal{L}_{gd}$}
\end{subfigure}
\begin{subfigure}[b]{\wsc\linewidth}
    \centering
    \caption{$\mathcal{L}_{total}$}
\end{subfigure}
\caption{Qualitative comparison on geometry losses. }
\label{fig:exp_geo}
\end{figure}

\subsection{Quantitative Results}
We carried out metric evaluations on three style translations, i.e., daytime to golden hour, daytime to blue hour, and daytime to nighttime, and reported mean scores in Tables \ref{table:exp_metrics} and \ref{table:exp_metrics_opt}. Our translation results and optimized results (Ours and Ours-opt) outperform others or have competitive performance on style transfer accuracy and quality (Accuracy, IS, IoU), structure and perceptual similarity (edge-SSIM, IoU). It validates the effectiveness of geometry losses, and our blending optimization (Ours-opt) can somehow restore primal scene contour information (edge-SSIM) and improve perceptual similarity (IoU). Overall, our models have better style transfer in terms of photorealism and style diversity.

\begin{table}[!t]
\centering
\renewcommand*{\arraystretch}{1.1}
\begin{tabular}{l | c  H  c  c  c}
\toprule
  & e-SSIM$\uparrow$ & LPIPS$\downarrow$ & Acc$\uparrow$ & IS$\uparrow$ & IoU$\uparrow$ \\
\midrule
Ours-whole & 0.6838 & 0.2350 & 0.8282 & 2.5240 & 0.7410 \\
Ours & 0.6359 & 0.2892 & 0.9486 & 2.7290 & 0.7257 \\
Ours-opt & 0.8094 & LPIPS & 0.9007 & 2.6127 & 0.7715 \\
\bottomrule
\end{tabular}
\caption{Ablation study of segmentation. Ours-whole indicates our full model trained with whole images, while Ours (or Ours-opt) is the full model trained with segmented images.}
\label{table:exp_metrics_segmentation}
\end{table}

\begin{table}[!t]
\centering
\renewcommand*{\arraystretch}{1.1}
\begin{tabular}{l | c  c  c | c  }
\toprule
  & w/o $\mathcal{L}_{kl} + \mathcal{L}_{gd}$ & w/o $\mathcal{L}_{kl}$ & w/o $\mathcal{L}_{gd}$ & $\mathcal{L}_{total}$ \\
\midrule
e-SSIM$\uparrow$  & 0.4800 & 0.5539 & 0.5159 & \textbf{0.6359}  \\
Acc$\uparrow$ & 0.8934 & 0.9201 & 0.9265 & \textbf{0.9486} \\
IS$\uparrow$ & 2.6858 & 2.7183 & 2.7241 & \textbf{2.7290} \\
IoU$\uparrow$ & 0.6056 & 0.6536 & 0.6612 & \textbf{0.7257} \\
\bottomrule
\end{tabular}
\caption{Ablation study of different geometry losses for foreground models. \textbf{Bold} text indicates the best result.}
\label{table:exp_metrics_losses}
\end{table}

\begin{figure*}[!t]
\centering
\def\wscn{0.162}
\def\hscn{0.164}
\includegraphics[width=0.20\linewidth,height=0.11\linewidth]{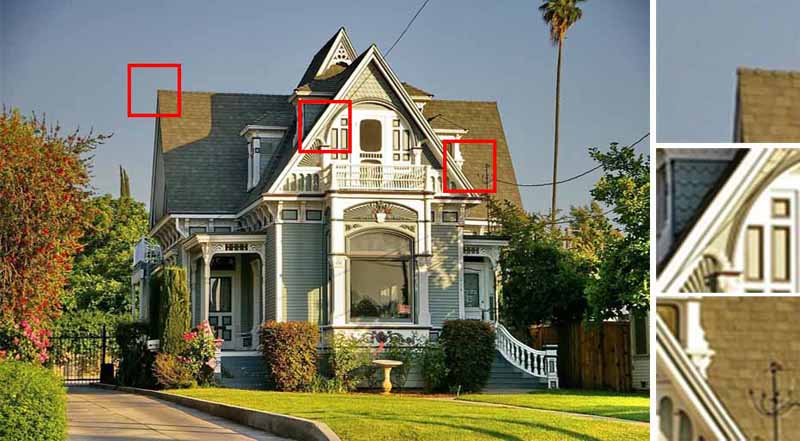}
\vspace{-0.2em}
\includegraphics[width=0.20\linewidth,height=0.11\linewidth]{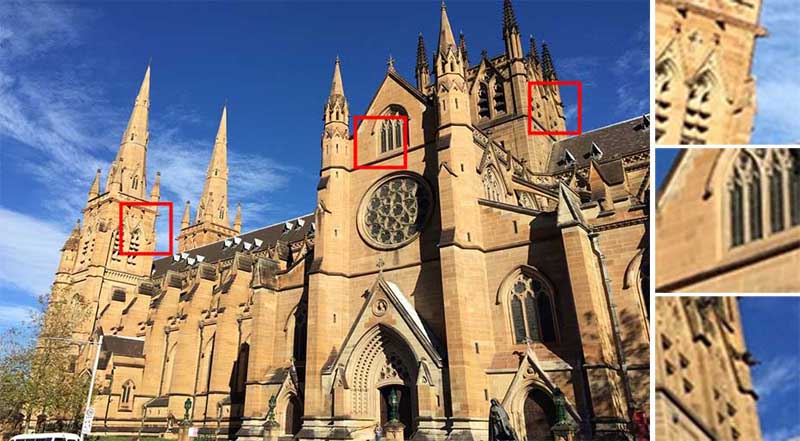}
\hspace{1.3em} 
\includegraphics[width=0.162\linewidth,height=0.11\linewidth]{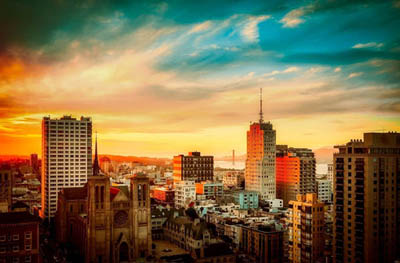}
\includegraphics[width=0.162\linewidth,height=0.11\linewidth]{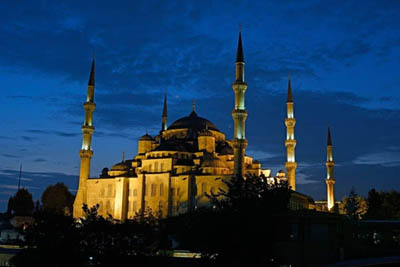}
\includegraphics[width=0.162\linewidth,height=0.11\linewidth]{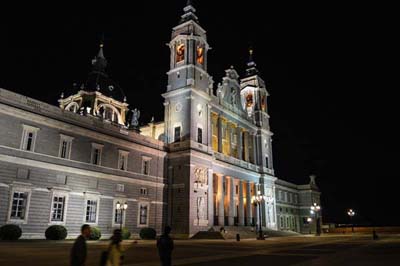}
\centerline{\hspace{1em} Original scenes \hspace{18em} Reference styles \hspace{2em} } 

\vspace{0.5em} 

\includegraphics[width=\wscn\linewidth,height=\hscn\linewidth]{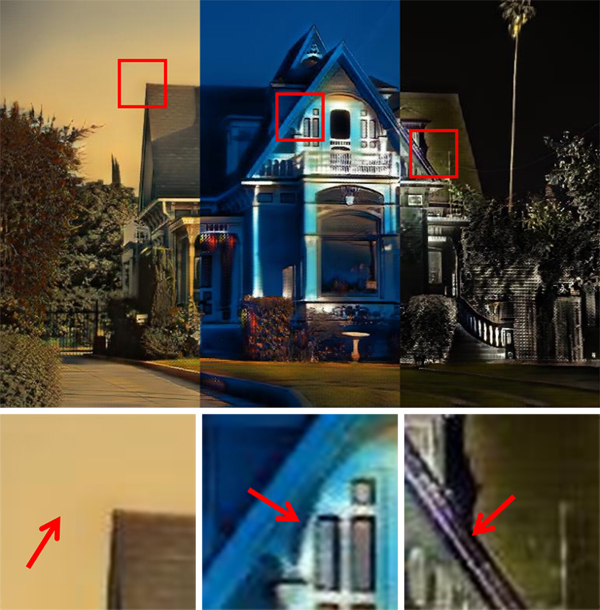}
\vspace{0.1em} 
\includegraphics[width=\wscn\linewidth,height=\hscn\linewidth]{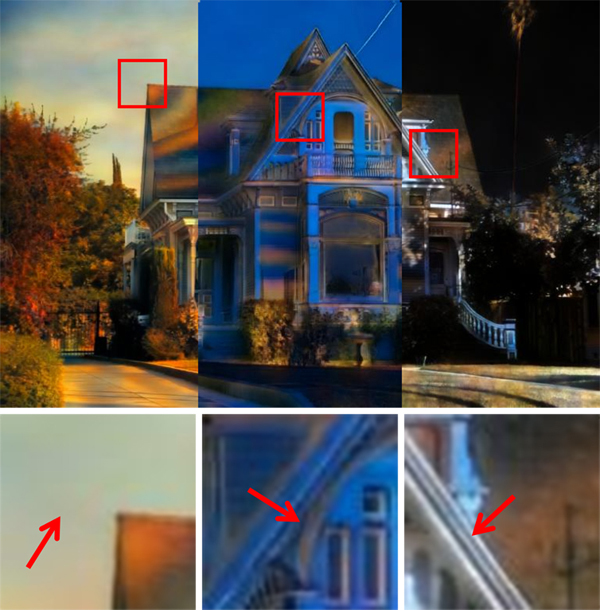}
\includegraphics[width=\wscn\linewidth,height=\hscn\linewidth]{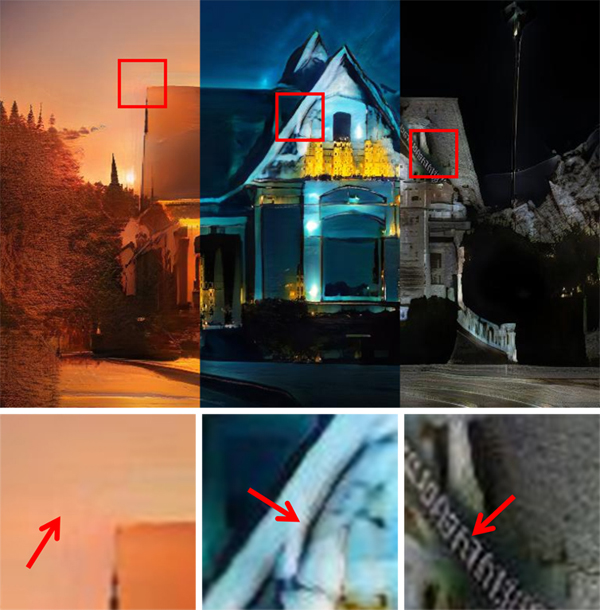}
\includegraphics[width=\wscn\linewidth,height=\hscn\linewidth]{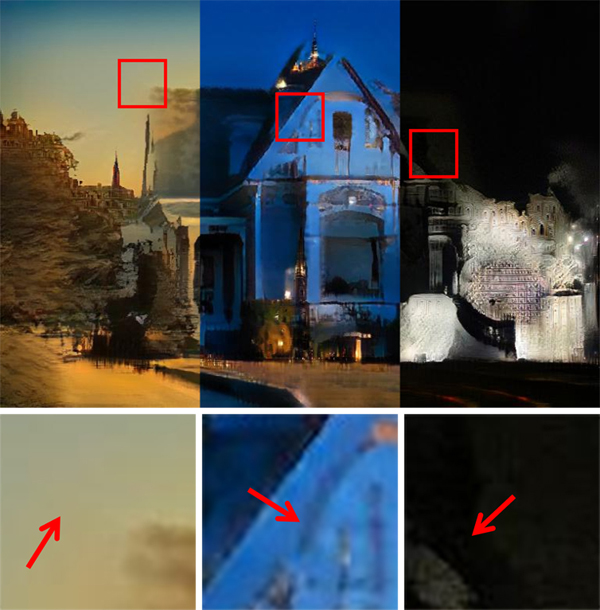}
\includegraphics[width=\wscn\linewidth,height=\hscn\linewidth]{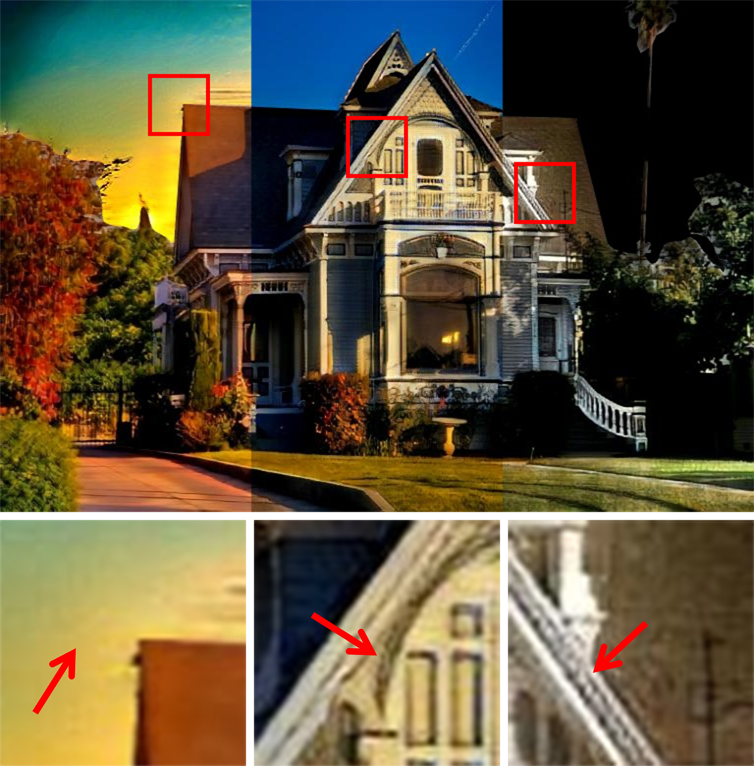}
\includegraphics[width=\wscn\linewidth,height=\hscn\linewidth]{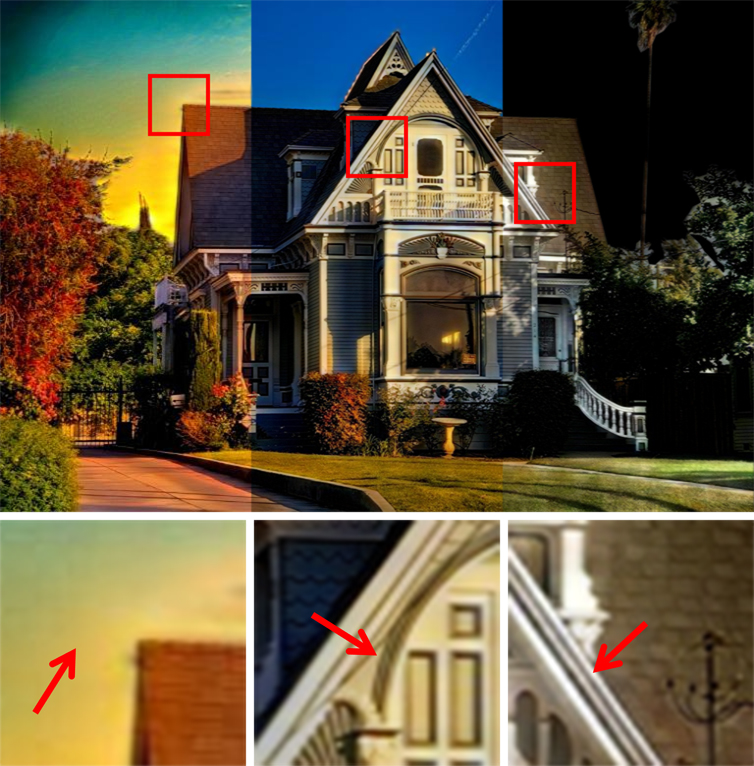}

\vspace{0.5em} 
\includegraphics[width=\wscn\linewidth,height=\hscn\linewidth]{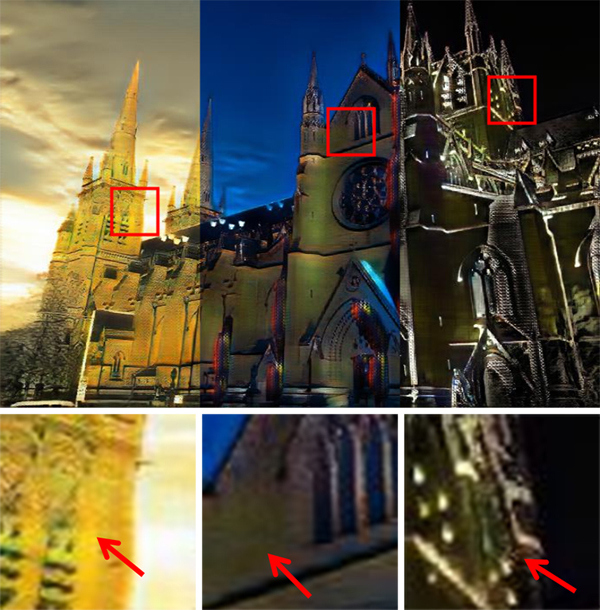}
\vspace{0.1em} 
\includegraphics[width=\wscn\linewidth,height=\hscn\linewidth]{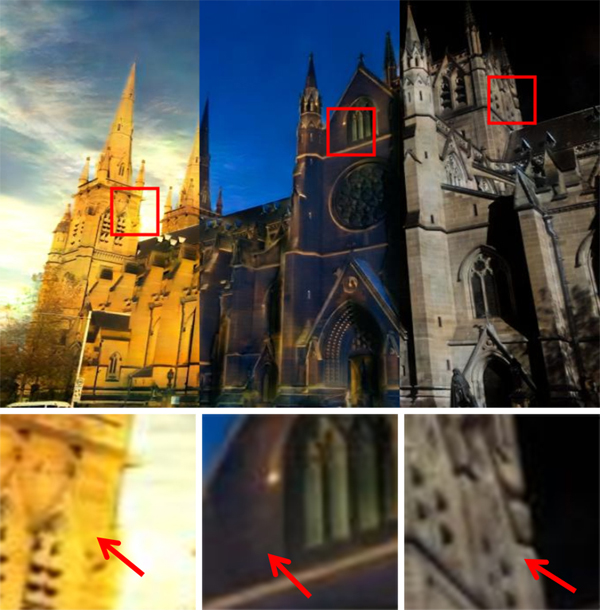}
\includegraphics[width=\wscn\linewidth,height=\hscn\linewidth]{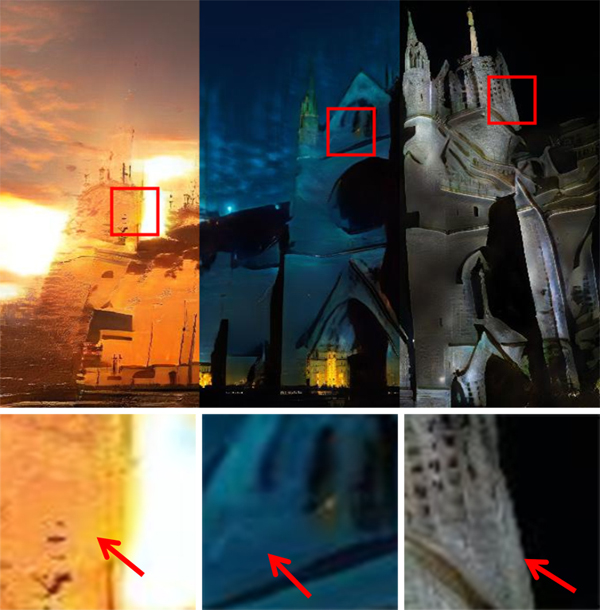}
\includegraphics[width=\wscn\linewidth,height=\hscn\linewidth]{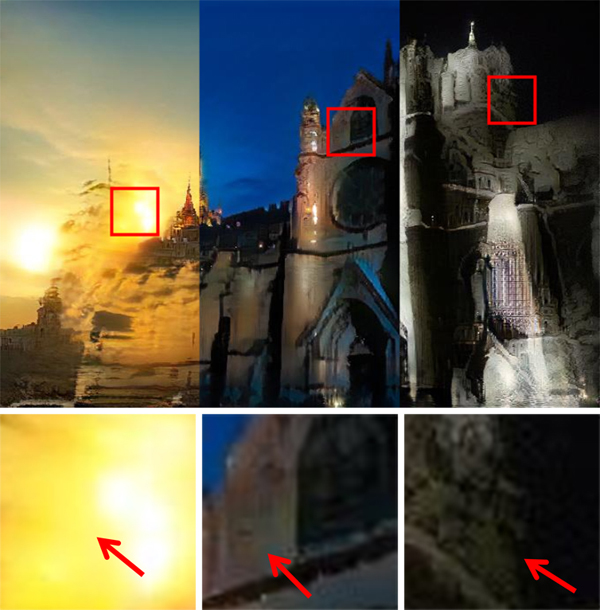}
\includegraphics[width=\wscn\linewidth,height=\hscn\linewidth]{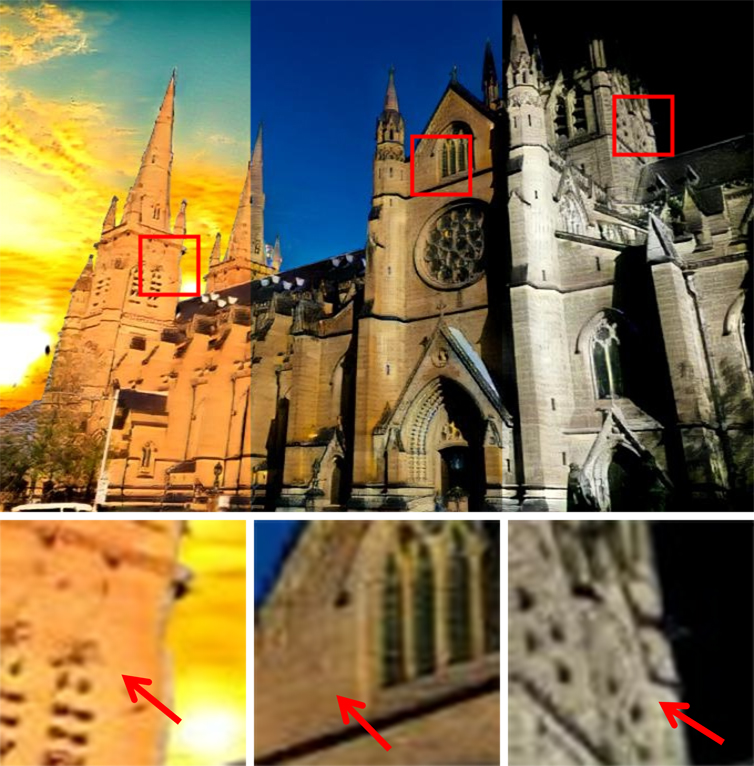}
\includegraphics[width=\wscn\linewidth,height=\hscn\linewidth]{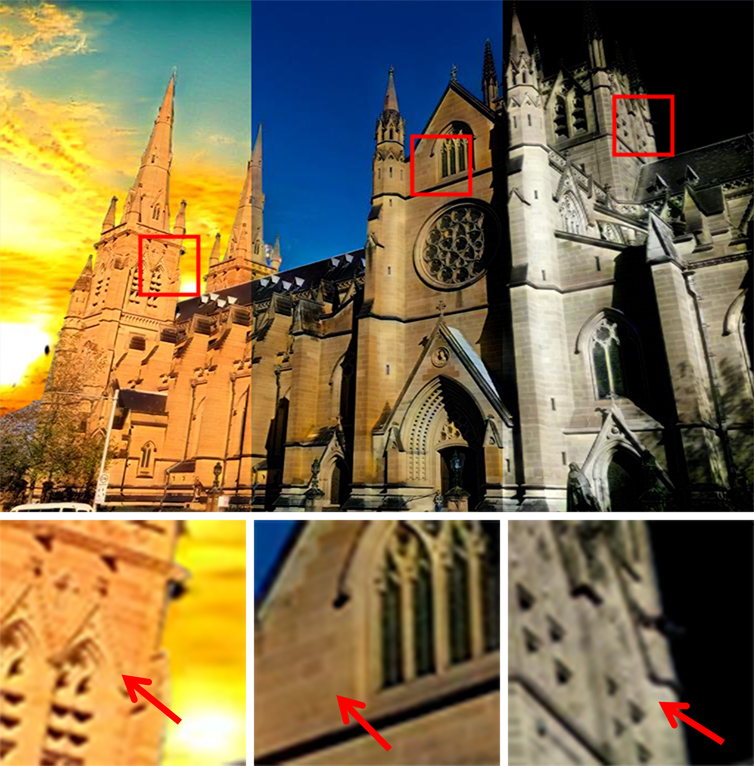}
\centerline{\hspace{0.5em} DRIT++\cite{lee2018diverse,lee2020drit++} \hspace{3em} MUNIT\cite{huang2018multimodal} \hspace{4em} FUNIT\cite{liu2019few}  \hspace{3.5em} DSMAP\cite{chang2020domain} \hspace{4.5em} Ours \hspace{5em} Ours+Opt \hspace{2em}}

\caption{Comparisons among image-to-image translation baselines and our proposed method. Our results have plausible colors from foreground
and background, and preserve the geometry in different style transfer cases. 
(Please see our interactive viewer in the supplementary for a detailed comparison.)
}
\label{fig:exp_qual_baselines}
\end{figure*} 

\begin{figure*}[!t]
\centering
\def\wscn{0.162}
\def\hscn{0.164}
\includegraphics[width=0.20\linewidth,height=0.11\linewidth]{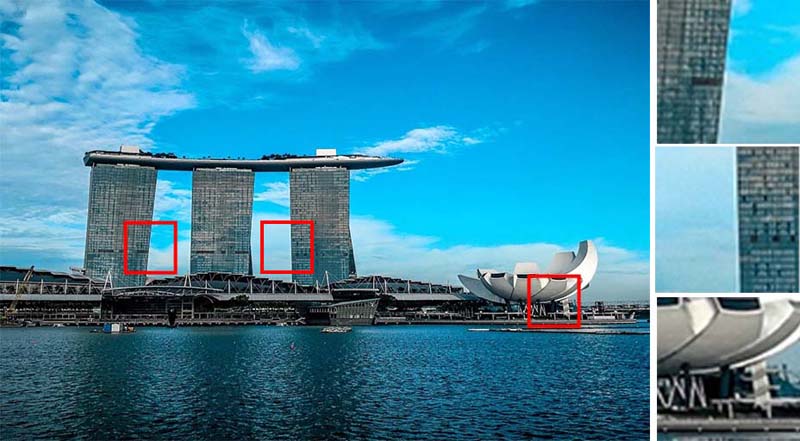}
\vspace{-0.2em}
\includegraphics[width=0.20\linewidth,height=0.11\linewidth]{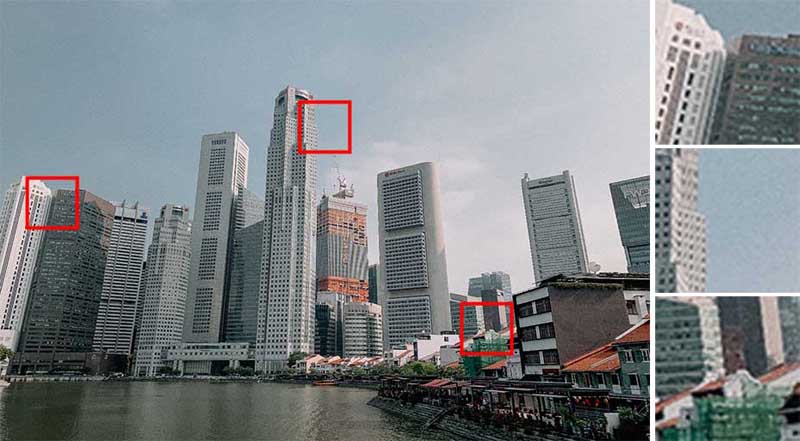}
\hspace{1.3em} 
\includegraphics[width=0.162\linewidth,height=0.11\linewidth]{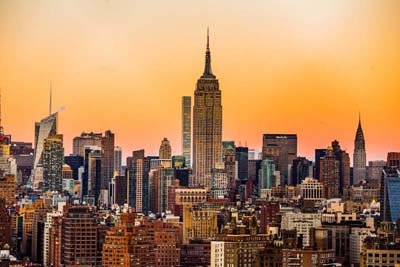}
\includegraphics[width=0.162\linewidth,height=0.11\linewidth]{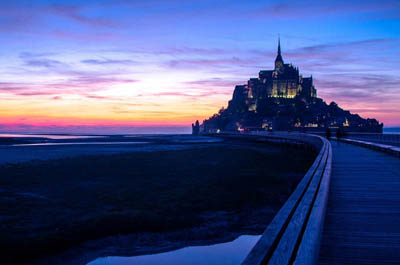}
\includegraphics[width=0.162\linewidth,height=0.11\linewidth]{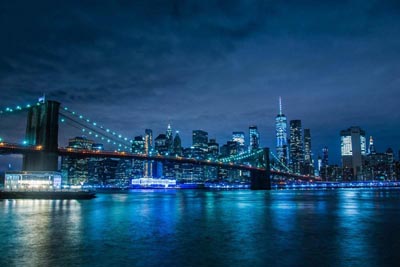}
\centerline{\hspace{0.5em} Original scenes \hspace{18em} Reference styles \hspace{2em} } 

\vspace{0.5em} 

\includegraphics[width=\wscn\linewidth,height=\hscn\linewidth]{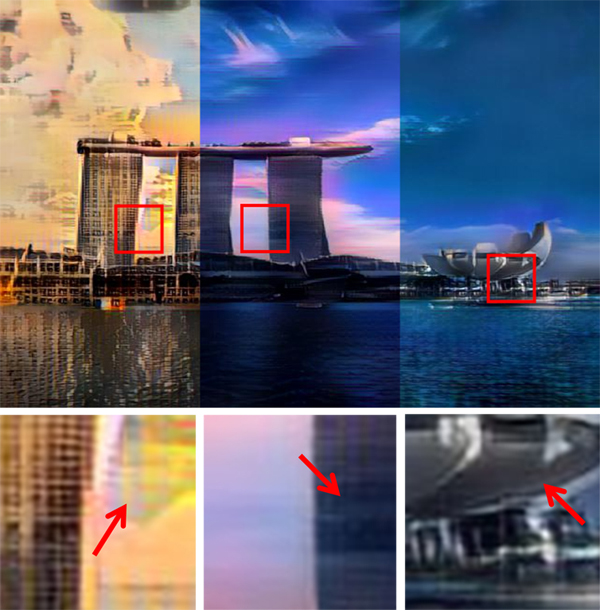}
\vspace{0.1em} 
\includegraphics[width=\wscn\linewidth,height=\hscn\linewidth]{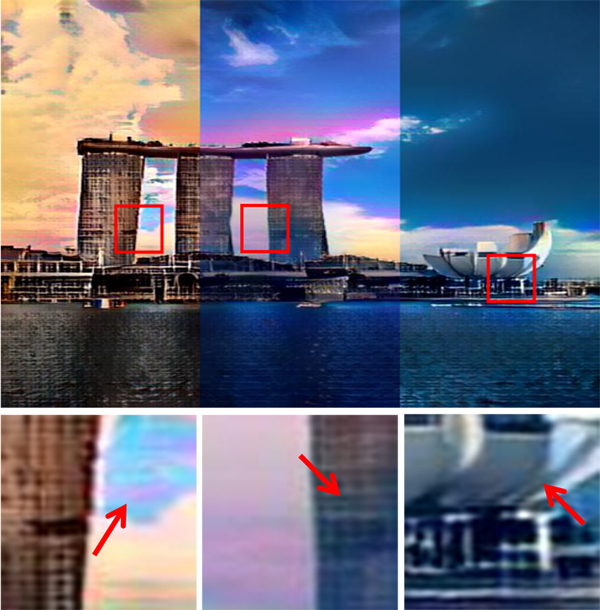}
\includegraphics[width=\wscn\linewidth,height=\hscn\linewidth]{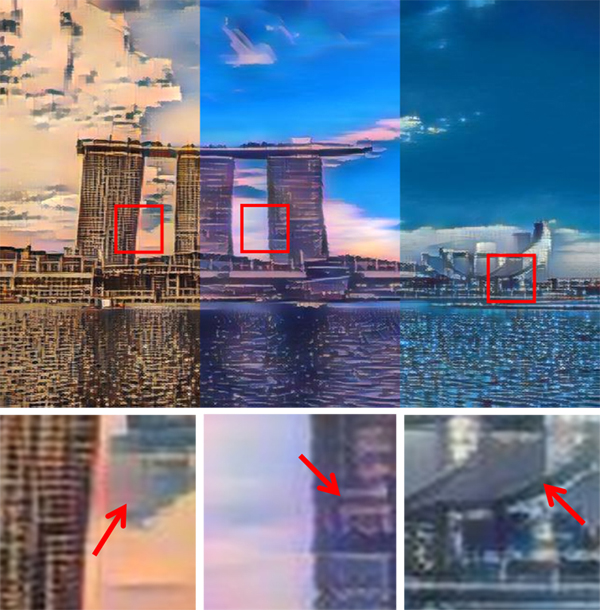}
\includegraphics[width=\wscn\linewidth,height=\hscn\linewidth]{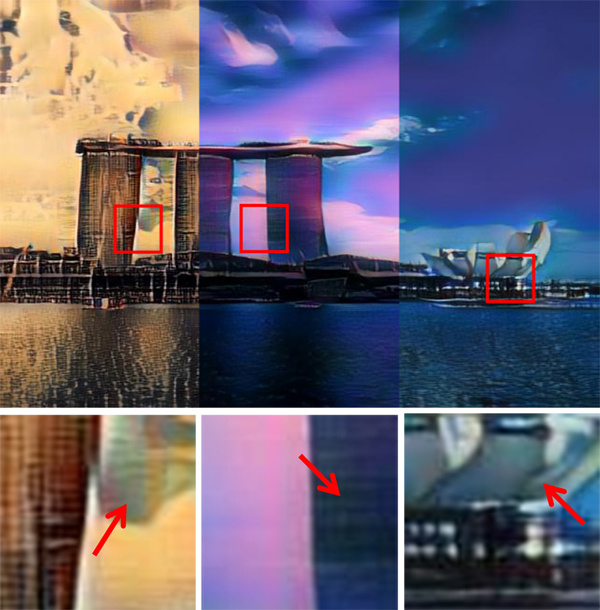}
\includegraphics[width=\wscn\linewidth,height=\hscn\linewidth]{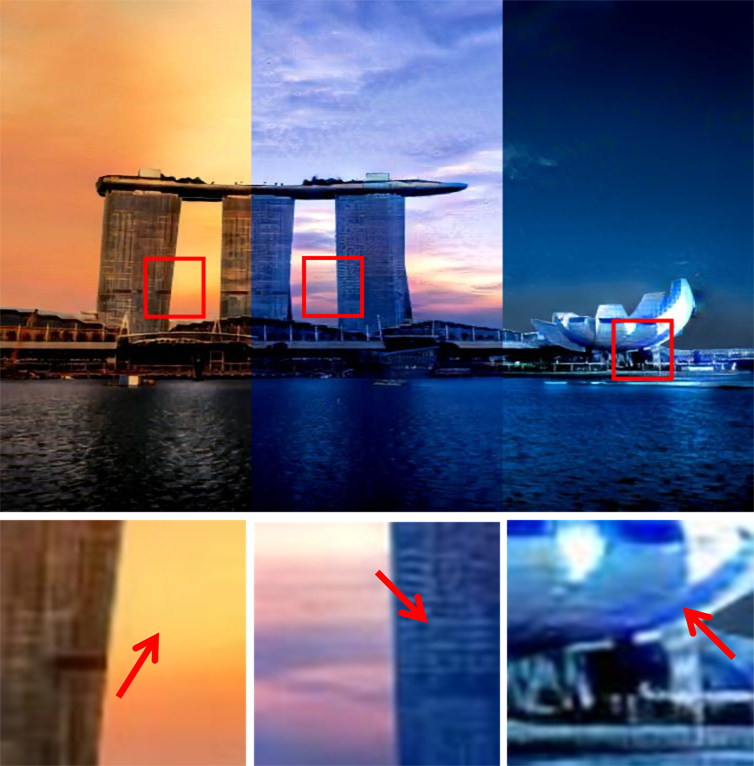}
\includegraphics[width=\wscn\linewidth,height=\hscn\linewidth]{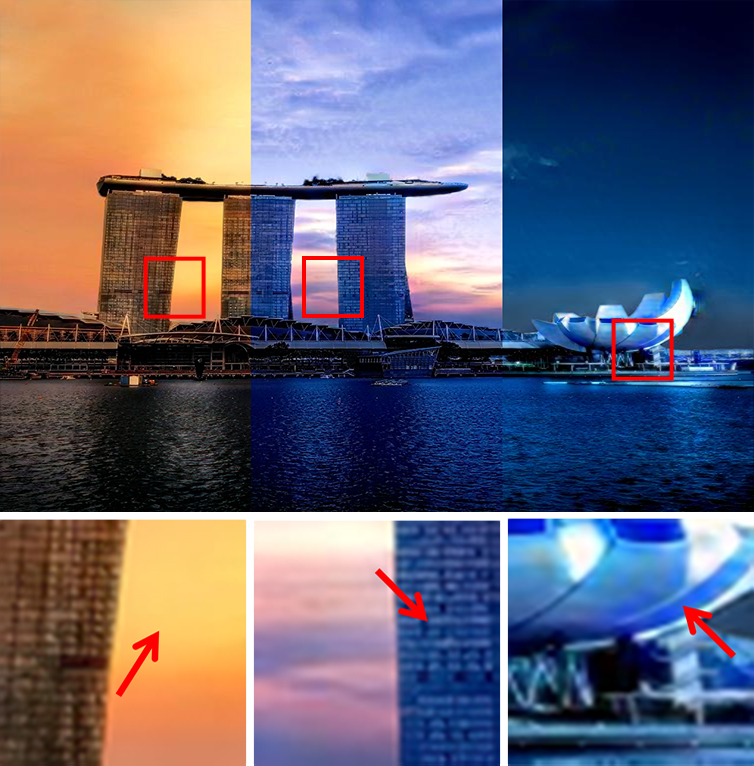}

\vspace{0.5em} 

\includegraphics[width=\wscn\linewidth,height=\hscn\linewidth]{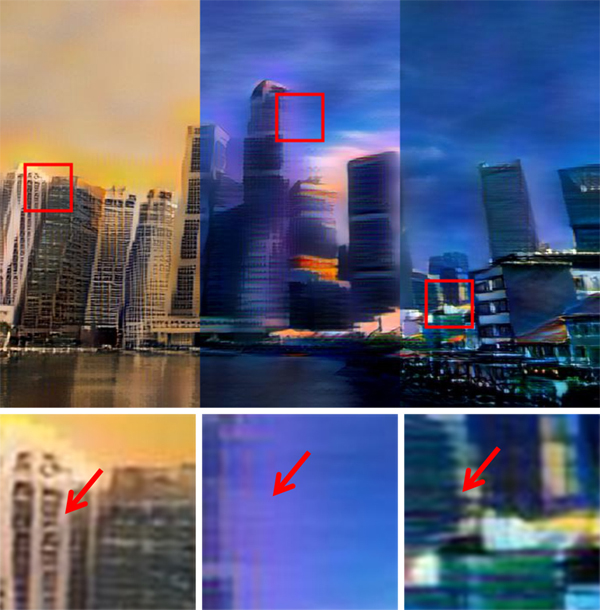}
\vspace{0.1em} 
\includegraphics[width=\wscn\linewidth,height=\hscn\linewidth]{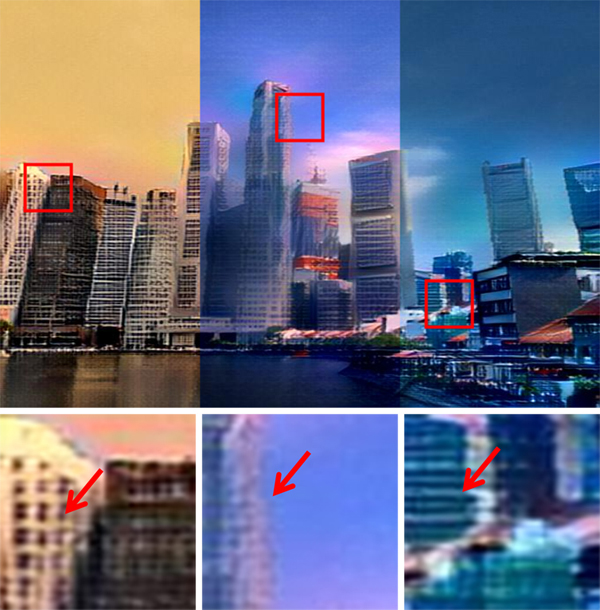}
\includegraphics[width=\wscn\linewidth,height=\hscn\linewidth]{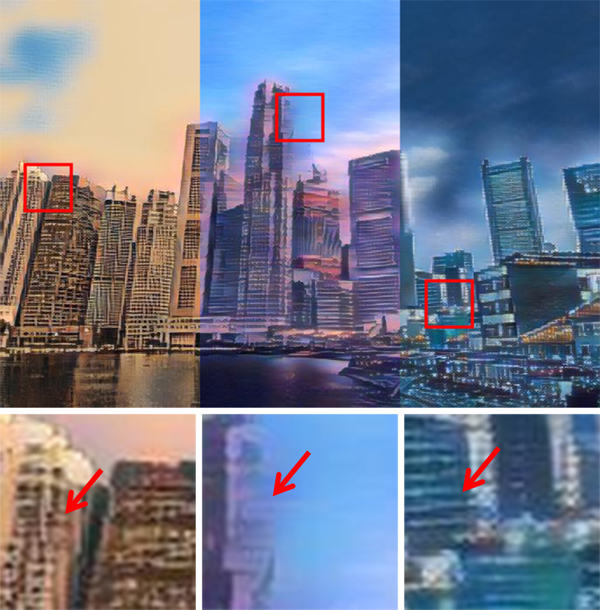}
\includegraphics[width=\wscn\linewidth,height=\hscn\linewidth]{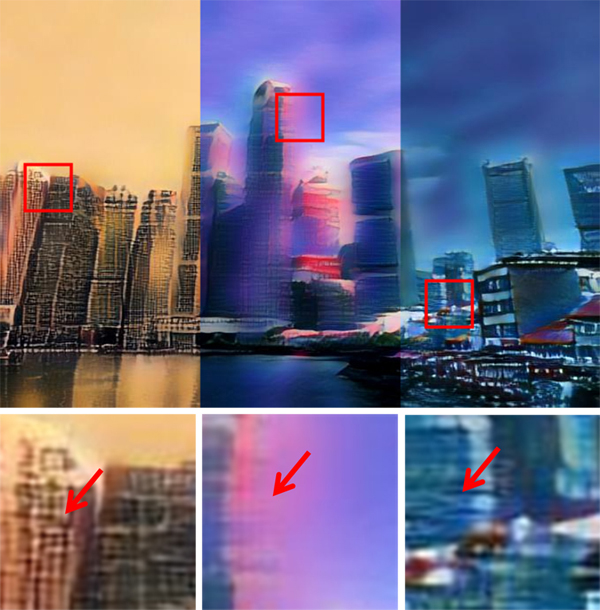}
\includegraphics[width=\wscn\linewidth,height=\hscn\linewidth]{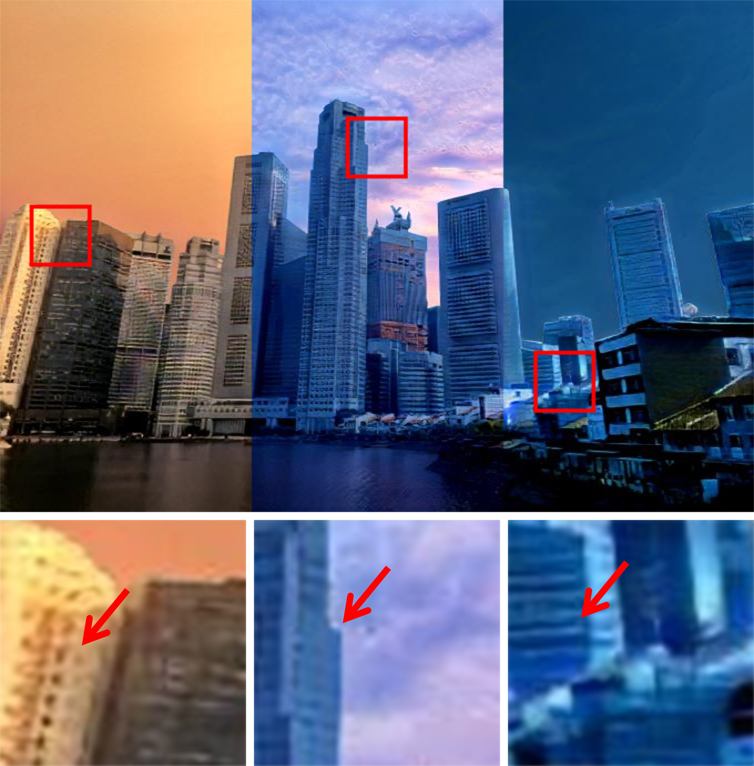}
\includegraphics[width=\wscn\linewidth,height=\hscn\linewidth]{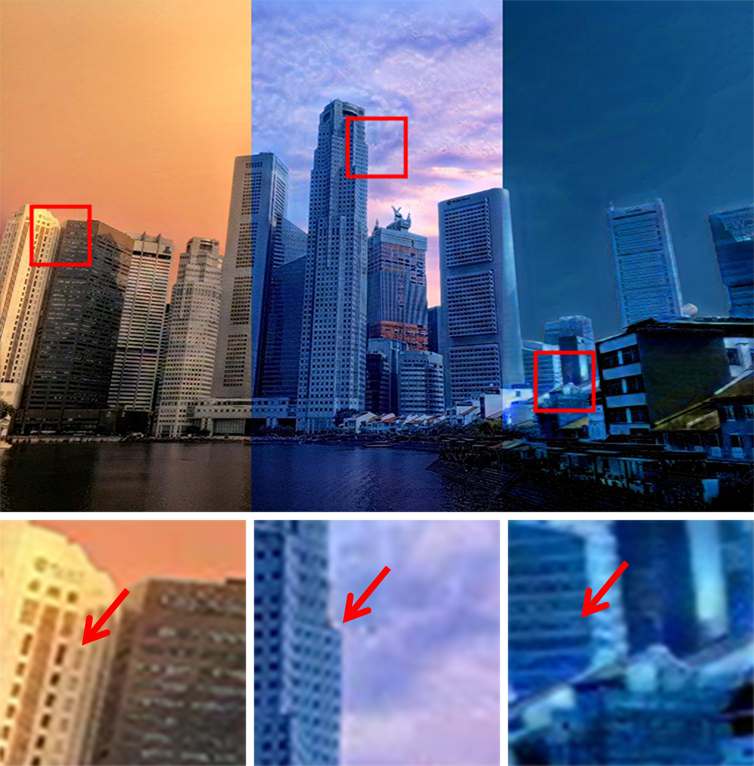}

\centerline{\hspace{1em}  AdaIN\cite{huang2017arbitrary} \hspace{4em} SANet\cite{park2019arbitrary} \hspace{3em} AdaAttN\cite{liu2021adaattn} \hspace{3.5em} LST\cite{li2019learning} \hspace{5.5em} Ours \hspace{5em} Ours+Opt \hspace{1.5em}}

\caption{Comparisons among neural style transfer baselines and proposed method. While neural style transfer methods tend to have visual artifacts, our results have matched colors from foreground
and background respectively, and preserve the geometry of the foreground while generating diverse cloud textures in the background. 
(Please see our interactive viewer in the supplementary for a detailed comparison.)
}
\label{fig:exp_qual_baselines_nst}
\end{figure*}

\subsection*{Ablation Study}

\noindent\textbf{Does segmentation help?}

We conducted ablation experiments on models trained with and without segmentation. On the one hand, segmentation provides explicit semantic color distribution information. As seen in Figure~\ref{fig:exp_seg}, some cases in which the model trained with whole images cannot semantically transfer color from buildings and sky, respectively. Our trained model using segmented foreground and background can successfully convey the correct color style semantically. On the other hand, with different hyperparameters of geometry losses for foreground and background models, our result Fig.~\ref{fig:exp_seg}(d) generates a dense cloud texture similar to style image Fig.~\ref{fig:exp_seg}(b) and keeps foreground appearance unchanged. But the result in Fig.~\ref{fig:exp_seg}(c) preserves both unchanged foreground and background texture and geometry.

\begin{figure}[t]
\centering
\includegraphics[width=0.48\textwidth]{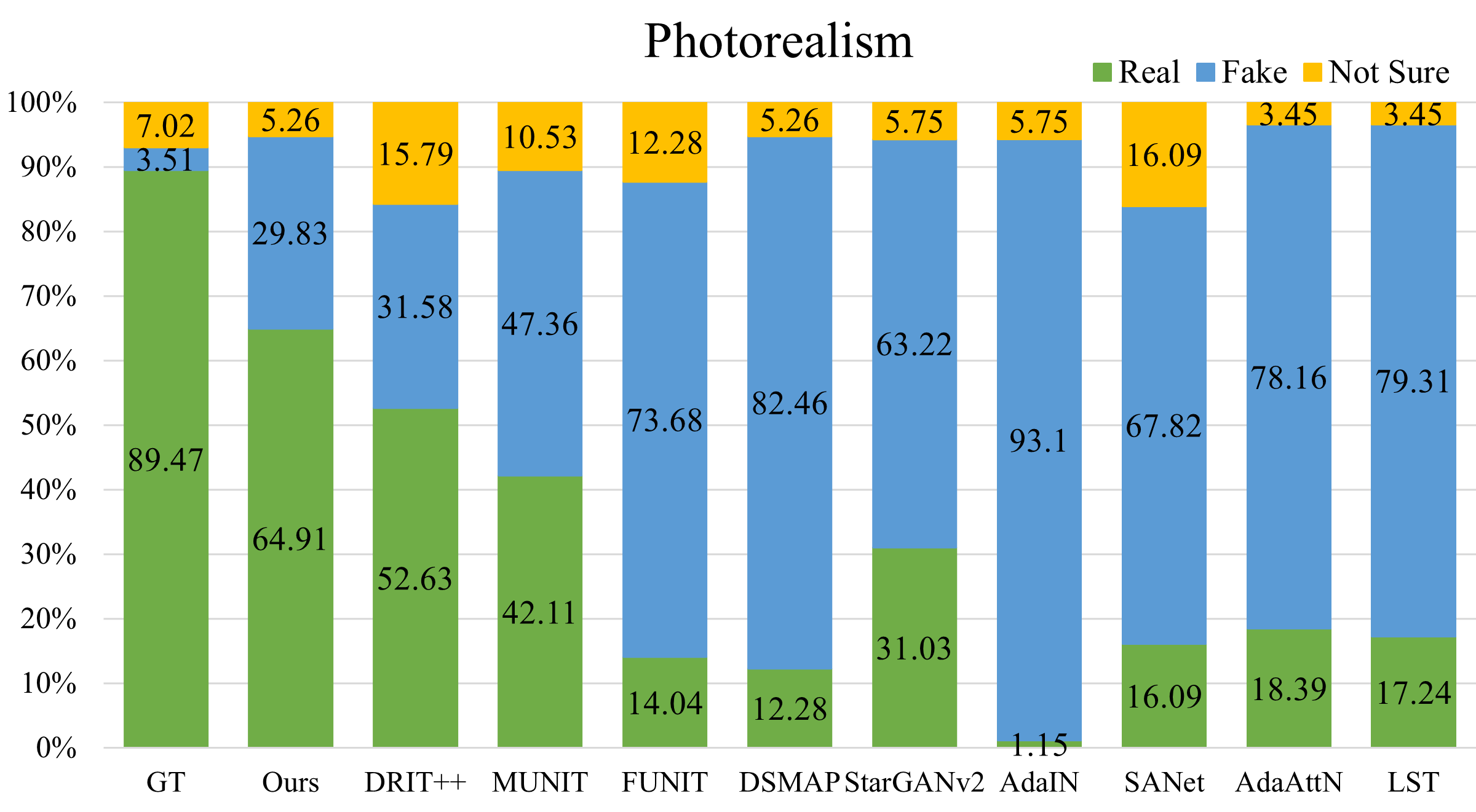}
\includegraphics[width=0.48\textwidth]{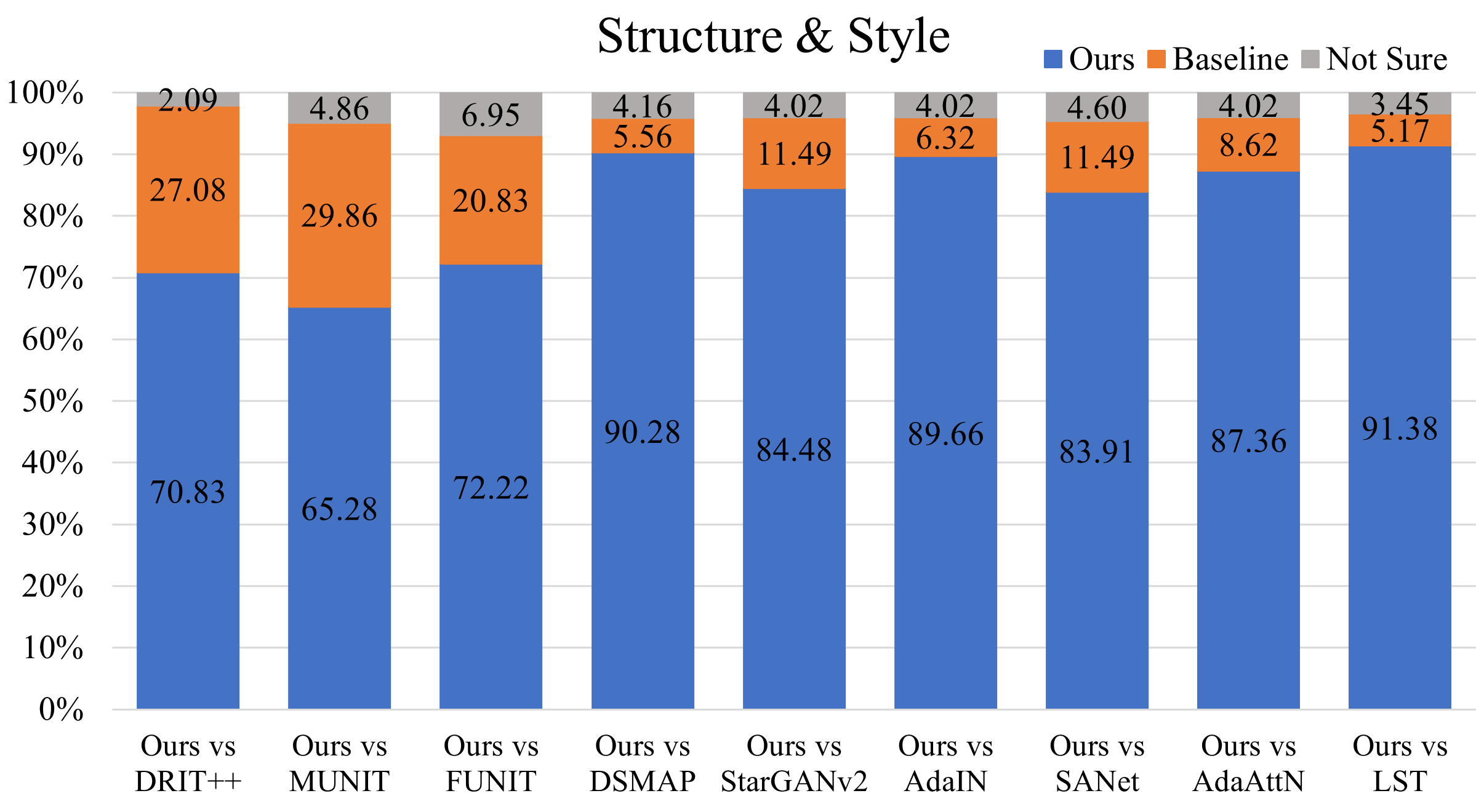}
\caption{Perceptual preference in terms of photorealism, structure and style.}
\label{fig:exp_user_study}
\end{figure}

The evaluated metric results are illustrated in Table \ref{table:exp_metrics_segmentation}. Compared to \textit{Ours}, \textit{Ours-whole} has better geometry preservation (e-SSIM, IoU) but worse stylization performance (Accuracy, IS). It might be because \textit{Ours} has different background from source input while \textit{Ours-whole} keeps whole image geometry information intact. The evaluation results in Table~\ref{table:exp_metrics} (IoU) and Figure~\ref{fig:exp_user_study} (Structure \& Style) also show that our segmented training strategy has better performance than baselines with respect to semantic style transfer.

\noindent\textbf{Are the geometry losses effective?} 

To validate the proposed geometry losses (\textit{Image Gradient loss} and \textit{Spatial Luminance KL Divergence loss}), we train foreground models with different settings: no geometry loss, without KL divergence loss, without image gradient loss, and with full geometry losses. 

According to qualitative results in Figure \ref{fig:exp_geo} and quantitative results in Table \ref{table:exp_metrics_losses}, a model with either only KL divergence loss or only gradient loss can to some extent help preserve geometry but does not perform better than a model with both geometry losses. Some previous work \cite{mechrez2017photorealistic} tried applying gradient loss in style transfer network but failed. We conjecture that the generator attains insufficient geometry transfer information from only gradient loss, e.g., with unconstrained luminance distribution, so infinite geometry changes have similar or even equal gradient losses. 
With the limit of spatial luminance distribution ($\mathcal{L}_{kl}$), \revised{gradient loss can protect primal geometry with geometry luminance density preservation while transferring sufficient style from source domain to target domain}.
In general, gradient loss plus spatial luminance KL divergence loss can largely improve the correctness of \revised{geometry} luminance transfer and keep geometry unchanged.

\subsection*{User Study}

To validate our results, we also conducted a perceptual user study (Fig.~\ref{fig:exp_user_study}) covering three aspects, i.e., image photorealism and semantic structure similarity plus semantic style consistency. The photorealism score contains the percentage of images that look real or fake. Structure and style score is obtained from pair comparisons (ours versus other baselines). Semantic structure similarity illustrates how well generated images keep foreground geometry intact and transfer target background texture. Semantic style consistency shows how much style transfers correctly for foreground and background. The results in Fig.~\ref{fig:exp_user_study} show that the proposed method outperforms previous works in terms of image fidelity and semantic style matching, indicating that our method can achieve more photorealistic style transfer than others. Particularly, neural style transfer methods generate much more non-photorealistic images than most image-to-image translation approaches.
Please refer to supplementary materials for the complete quantitative results and the user study details.

\subsection{Qualitative Results} \label{exp_quality}

\begin{figure}[!t]
\centering
\def\sc{0.30}
\def\scb{0.2}
\includegraphics[width=\sc\linewidth]{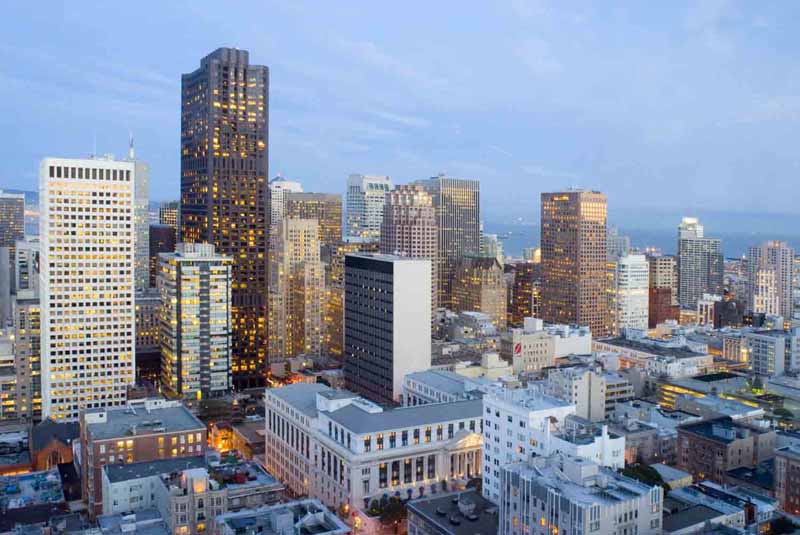}
\vspace{-0.3\baselineskip} 
\includegraphics[width=\sc\linewidth]{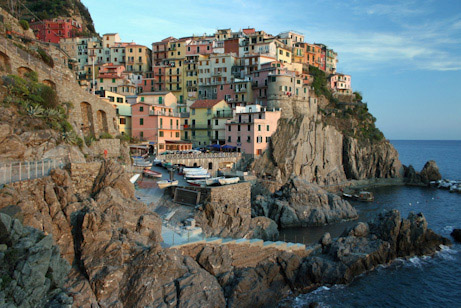}
\includegraphics[width=\sc\linewidth,height=\scb\linewidth]{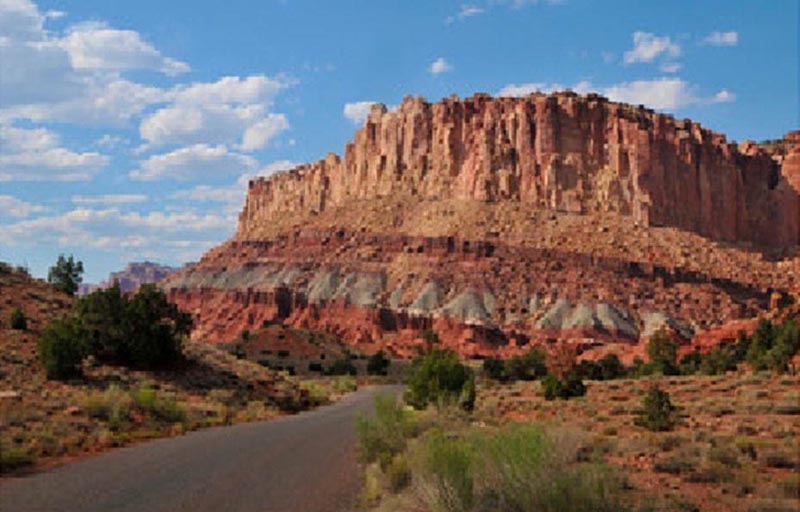}

\centerline{Input}
\vspace{0.1\baselineskip}

\includegraphics[width=\sc\linewidth,height=\scb\linewidth]{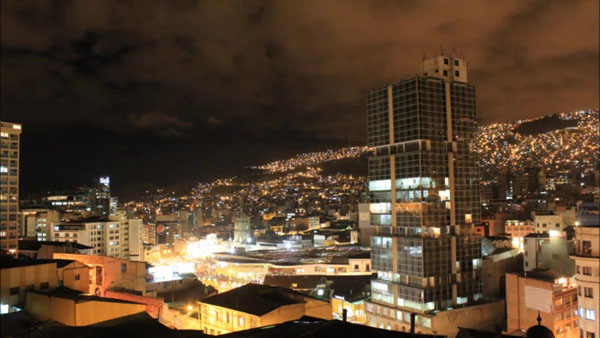}
\vspace{-0.3\baselineskip} 
\includegraphics[width=\sc\linewidth,height=\scb\linewidth]{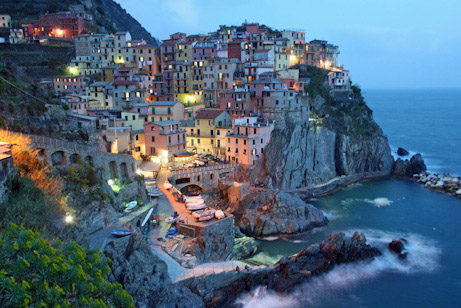}
\includegraphics[width=\sc\linewidth]{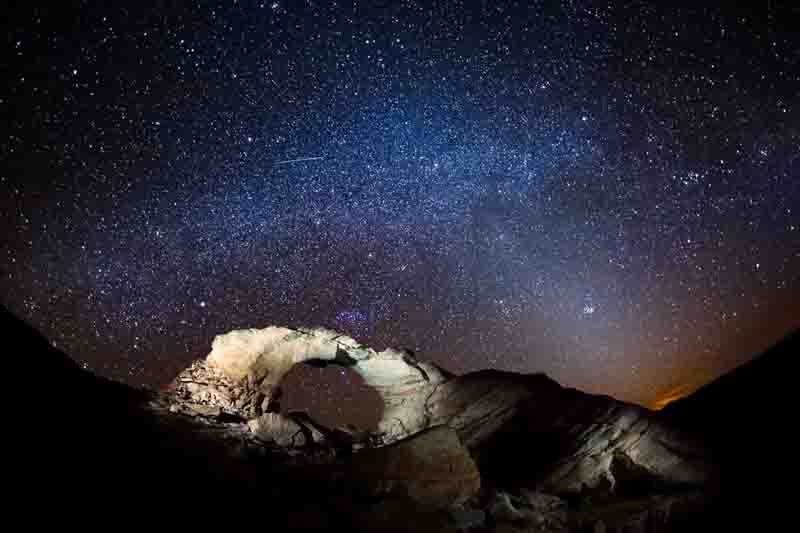}

\centerline{Reference styles}
\vspace{0.1\baselineskip}

\includegraphics[width=\sc\linewidth,height=\scb\linewidth]{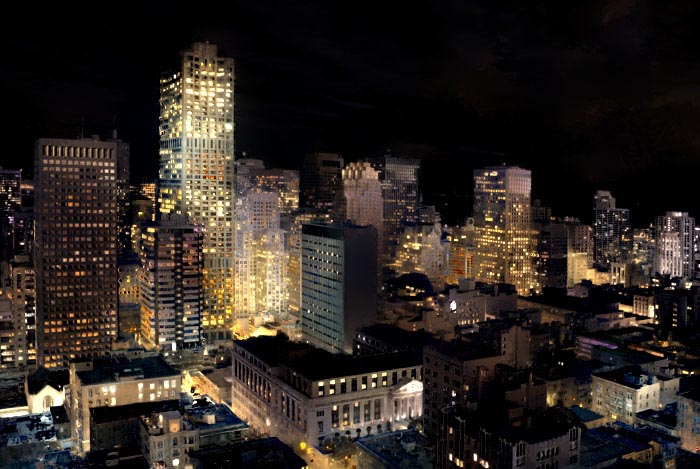}
\vspace{-0.3\baselineskip} 
\includegraphics[width=\sc\linewidth,height=\scb\linewidth]{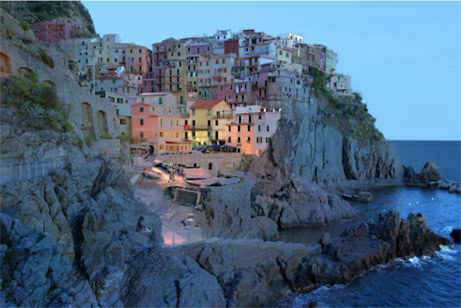}
\includegraphics[width=\sc\linewidth,height=\scb\linewidth]{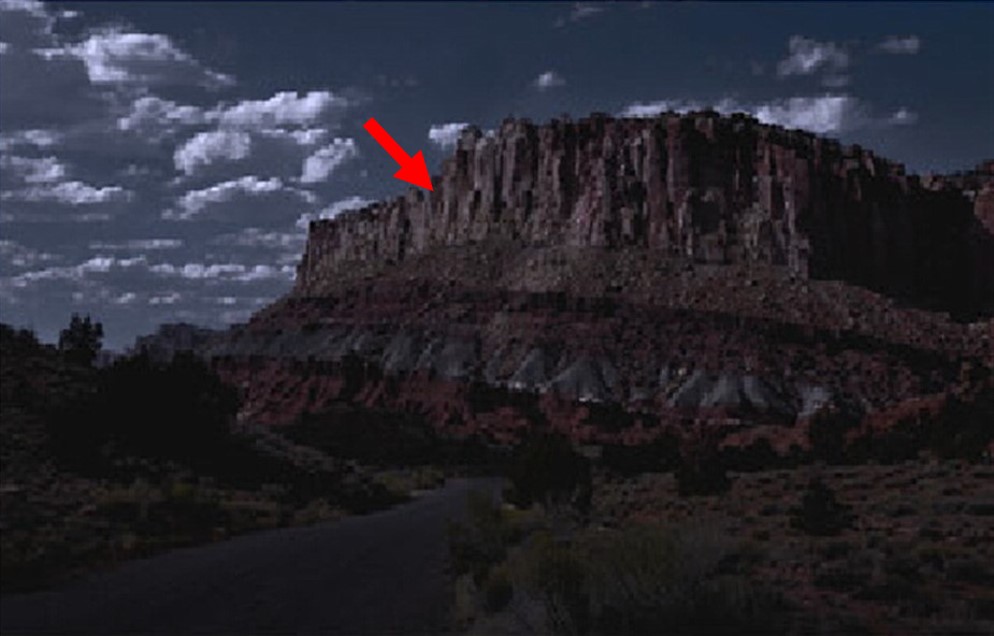}

\centerline{Shih et al.~\cite{shih2013data} \quad Laffont et al.~\cite{laffont2012coherent} \quad Pouli et al.~\cite{pouli2011progressive}}
\vspace{0.1\baselineskip}

\includegraphics[width=\sc\linewidth,height=\scb\linewidth]{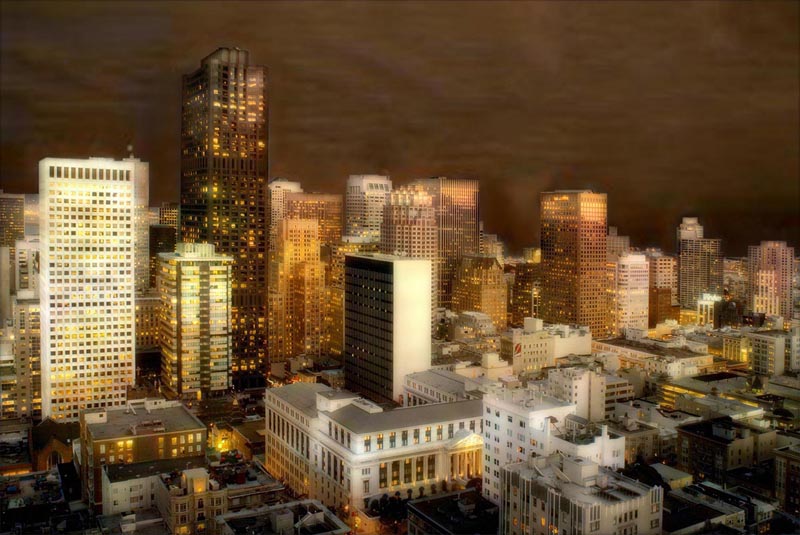}
\vspace{-0.3\baselineskip} 
\includegraphics[width=\sc\linewidth,height=\scb\linewidth]{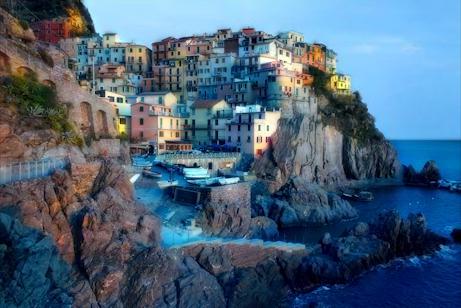}
\includegraphics[width=\sc\linewidth,height=\scb\linewidth]{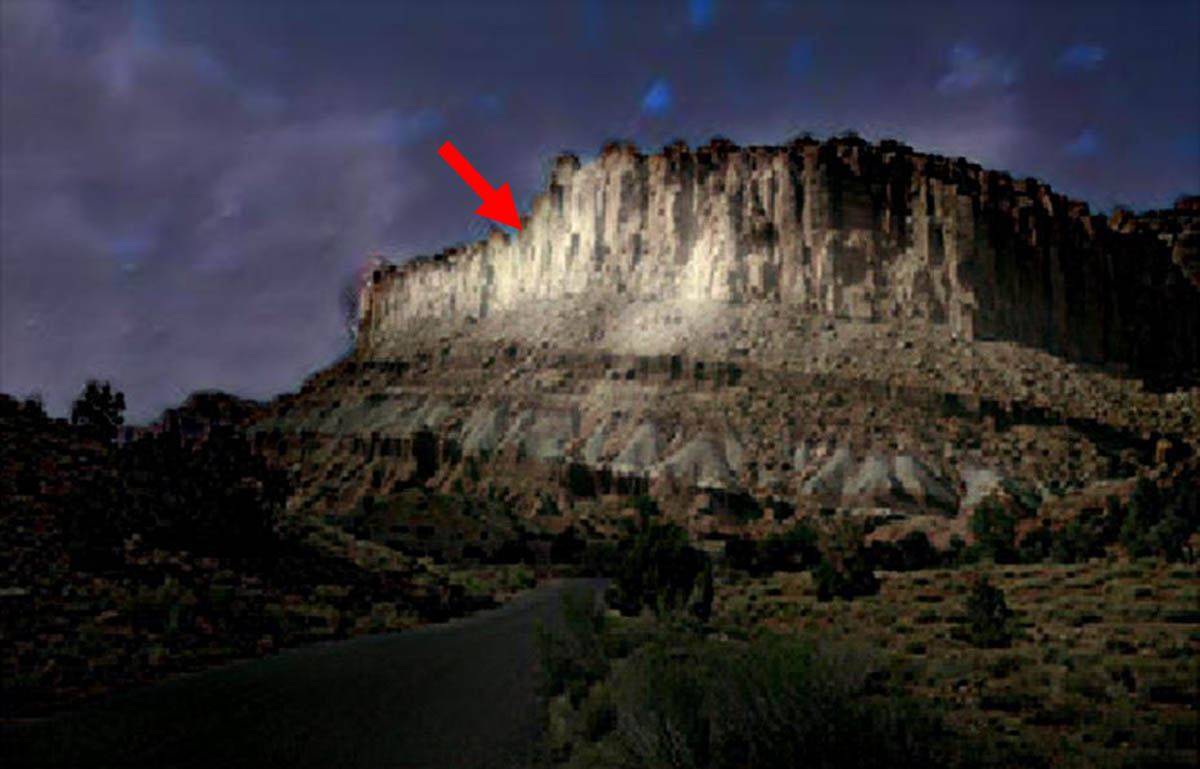}
\centerline{Ours}
\caption{Additional comparisons to traditional methods.}
\label{fig:additional_compare}
\end{figure} 

Results from some baselines for style transfers from daytime to golden hour, blue hour, and nighttime are selected to display in Figure \ref{fig:exp_qual_baselines} and Figure \ref{fig:exp_qual_baselines_nst}. More visual results can be checked in the supplementary.

In general, all baselines tend to have inaccurate semantic color matching. As can be seen, for golden or blue style transfer, all baselines treat some parts of the building as sky, so it leaks sky color (DRIT++, MUNIT, FUNIT, DSMAP, AdaIN,  SANet, AdaAttN) or texture (FUNIT, DSMAP, AdaIN, LST) to the foreground. 

FUNIT, and DSMAP tend to generate noisy artifacts, e.g., golden hour and nighttime style transfer. DSMAP and FUNIT sometimes merge part of the building into the background, hence appearing the ghosting artifacts and building distortion. By contrast, our approach can transfer a more semantically matched color style while retaining the foreground appearance. Sky in our results has both style and texture in line with the target style images, and our foreground has plausible new lighting style and geometry quality. 

From visual results in Fig.\ref{fig:exp_qual_baselines_nst}, neural style transfer methods (AdaIN, SANet, AdaAttN, and LST) in some way preserve global geometry because of the effect of content related losses. 
Nonetheless, these style transfer methods transfer texture strokes for both sky and foreground (e.g. sky in results in the first row, building texture in results in the second row). Ours tends to produce different sky textures according to the reference styles, such as removing clouds for golden style and adding clouds for blue style in Fig.\ref{fig:exp_qual_baselines_nst}. Particularly, AdaAttN has sharper appearance features since it constrains both local (high-frequency) and global (low-frequency) content features during training. But AdaattN fails to produce photorealistic appearance. By contrast, our proposed geometry losses are based on high-frequency spatial information, thus helping with geometry preservation. Regarding stylization performance, AdaIN, SANet, and LST have visually more aesthetic style effects than AdaAttN and globally matched color in terms of reference styles, though the transferred images look like paintings.
In general, our results are more photorealistic, while neural style transfer baselines are more artistic.

\subsection{Comparisons to Traditional Methods} \label{exp:trad_methods}
We provide additional comparisons to non-deep learning methods including color transfer by Shih \textit{et al.}\cite{shih2013data}, by Pouli and Reinhard \cite{pouli2011progressive}, and an intrinsic decomposition for relighting method by Laffont \textit{et al.}\cite{laffont2012coherent} in Figure~\ref{fig:additional_compare}.
Shih et al.'s locally affine model is solved from a pair of frames of a similar scene retrieved from their video database. The intrinsic decomposition-based method by Laffont et al. \cite{laffont2012coherent} requires paired images that capture a typical scene under varying lighting for illumination transfer.
Our method instead learns the transfer with unpaired images.

For night-style images, Shih \textit{et al.} fail to handle the background properly, while our result has a natural sky. Our method also has more consistent styles with better color saturation than the method by Laffont \textit{et al.} and better highlights on rocks than the method by Pouli and Reinhard.


\begin{figure}[!t]
    \centering
    \small
    \def\sc{0.43}
    \begin{subfigure}[t]{\sc\linewidth}
        \centering
        \includegraphics[width=1\linewidth]{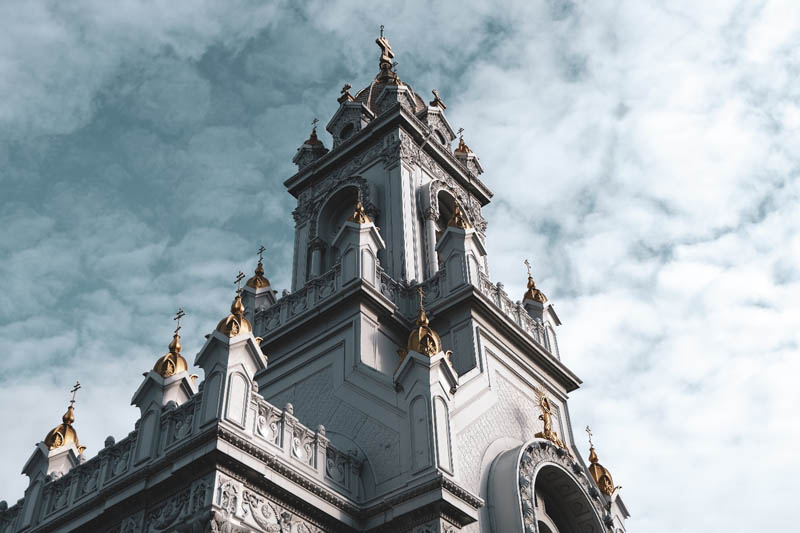}
        \vspace{-1.5\baselineskip} 
        \caption{Input}
    \end{subfigure}
    \begin{subfigure}[t]{\sc\linewidth}
        \centering
        \includegraphics[width=1\linewidth]{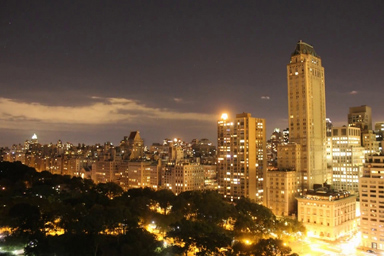}
        \vspace{-1.5\baselineskip} 
        \caption{Style}
    \end{subfigure}

    \vspace{0.1em}
    \begin{subfigure}[t]{\sc\linewidth}
        \centering
        \includegraphics[width=1\linewidth]{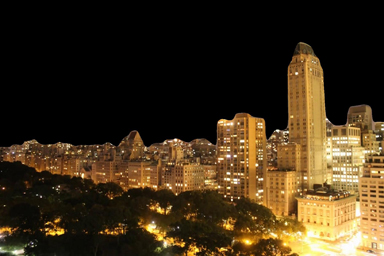}
        \vspace{-1.5\baselineskip} 
        \caption{Style foreground}
    \end{subfigure}
    \begin{subfigure}[t]{\sc\linewidth}
        \centering
        \includegraphics[width=1\linewidth]{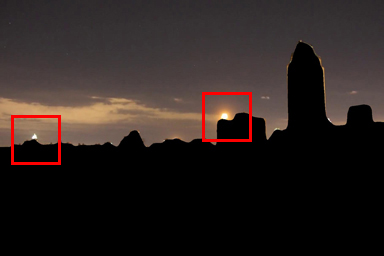}
        \vspace{-1.5\baselineskip} 
        \caption{Style background}
    \end{subfigure}
    
    \vspace{0.1em}
    \begin{subfigure}[t]{\sc\linewidth}
        \centering
        \includegraphics[width=1\linewidth]{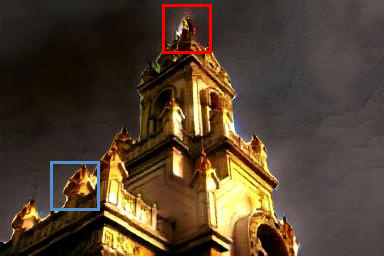}
        \vspace{-1.5\baselineskip} 
        \caption{Ours}
    \end{subfigure}
    \begin{subfigure}[t]{\sc\linewidth}
        \centering
        \includegraphics[width=1\linewidth]{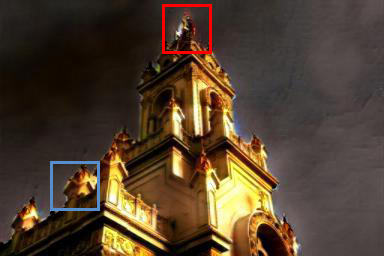}
        \vspace{-1.5\baselineskip} 
        \caption{Ours-opt}
    \end{subfigure}
    
    \caption{Failure cases. (Zoom in for a best view.)
    }
    \label{fig:fail_case}
\end{figure}

\subsection{Limitation}
\label{sec:limitation}
As our approach largely relies on segmentation, sometimes deficient segmentation (e.g., nighttime style background and daytime imperfect foreground in Fig.~\ref{fig:fail_case}) will lead to wrong color rendition or artifacts. For example, in Figure \ref{fig:fail_case}, the results generate shiny spot artifacts (red boxes in (e) and (f)) on the top of building due to wrong style background, and wrong bright yellow boundary (blue boxes in (e) and (f)) on background due to input foreground enclosing background portion.

\section{Conclusion} 

We realize a photorealistic neural style transfer system to solve the architectural style transfer problem, which transfers an outdoor architectural photograph from daytime style to a target style at different magical times in a day (i.e., golden hour, blue hour, and nighttime). An ideal architectural style transfer is able to achieve color and illuminance transfer for foreground buildings, roads, etc., and realize novel target sky background color and texture transfer. With foreground and background segmentation for training respectively and proposed geometry losses, our image-to-image translation models successfully transfer semantically matched styles and  effectively preserve content information of foreground.

Our proposed method is not limited to architectural photographs and can be generalized to transfer other types of images as long as they have foreground and background in different lighting and texture properties, e.g. headshot portrait transfer, group photos, natural landscape images with animals. 
Exploring these domains is interesting future work once sufficient data are collected.
Additionally, developing a more robust segmentation technique for complex outdoor scenes, and training the proposed networks in an end-to-end fashion would be interesting research avenue.

\ifpeerreview \else
\section*{Acknowledgments}
This paper was partially supported by an internal grant from HKUST (R9429) and the HKUST-WeBank Joint Lab.
\fi

\bibliographystyle{IEEEtran}
\bibliography{references}

\ifpeerreview \else

\fi

\clearpage

In the appendices, we first provide more details about the dataset and our image translation network implementation (Section \ref{sec:data_detail} and Section~\ref{sec:implement}).
Then, we provide perceptual user study details (Section \ref{sec:user_study}). Finally, we show more results (Section~\ref{sec:result}) including the complete quantitative results, and more visual results and comparisons in terms of style transfer and blending. Particularly, we display all baseline comparisons in an \href{https://chenyingshu.github.io/time_of_day/interactive_viewer}{interactive html viewer}.

\section*{Errata and Change log}
\revised{
\noindent\textbf{Oct 27, 2022}: During our code release after the paper was published, we found some inconsistencies between the description of the KL loss in the main text and the implementation in the code.
We therefore updated the manuscript to make it consistent with our implementation. 
The changes include Eq.~\ref{eq:lkl1}, Fig.~\ref{fig:framework},   Fig.~\ref{fig:translation_framework} and the relevant descriptions in \textbf{Spatial Luminance KL Divergence Loss} definition and \textbf{Ablation Study} of geometry losses.
We also corrected the statistics of the dataset in Table~\ref{tab:data_stat} in the appendix. Particularly, the number of \textit{``unpaired imgs``} for blue styles is changed from 3,567 to 2,567, the total \textit{``unpaired imgs``} from 16,908 to 15,908, and the total ``training set`` from 21,291 to 20,291.
}

\appendices
\section{Dataset Details} 
\label{sec:data_detail}

\subsection{Annotation Explanation} 
Here depicts the definition of the four labels of magic times of the day for the dataset (i.e., daytime, golden hours, blue hours and night time), following the same annotation settings as Shih \textit{et al.}'s work \cite{shih2013data}. 

Officially speaking, the golden hours occur just after sunrise over the horizon or before sunset when the sun falls closing to the horizon, creating the magical warm glow. The blue hours appear shortly before sunrise and after sunset with the sun’s position just below the horizon and produce the cooler tones. In addition to these four time slots, there are some hours worth photographing, such as civil, nautical, and astronomical hours \cite{marques}, which arrive between the blue hour and nighttime. To covers all enchanting hours in a day for style transfer, we simply merge blue, civil, nautical, and hours as blue moments, astronomical and nighttime as night hours.

\subsection{Dataset Statistics and Visual Examples}
Table \ref{tab:data_stat} illustrates the detailed statistics of our time-lapse architectural dataset. The evaluation set is a separate unseen set for the training set, consisting of high-fidelity real-world photos from public domains \cite{pexels,pikwizard, unsplash}. 

\begin{figure}[ht]
\centering
\includegraphics[width=0.24\linewidth]{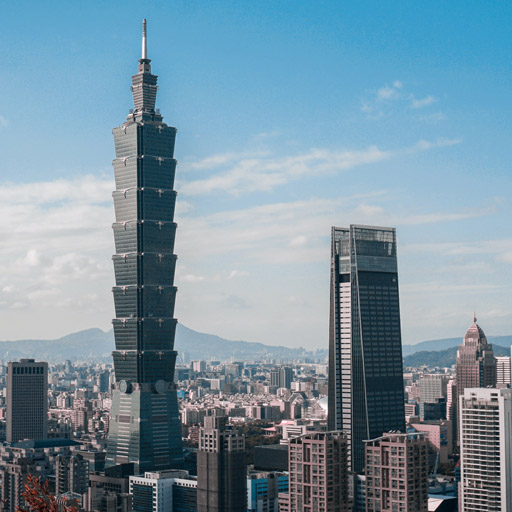}
\includegraphics[width=0.24\linewidth]{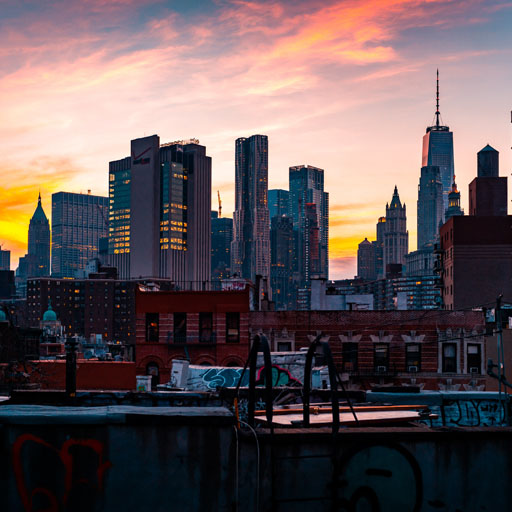}
\includegraphics[width=0.24\linewidth]{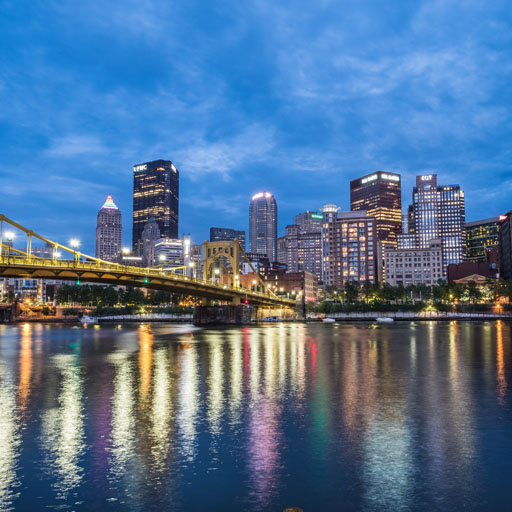}
\includegraphics[width=0.24\linewidth]{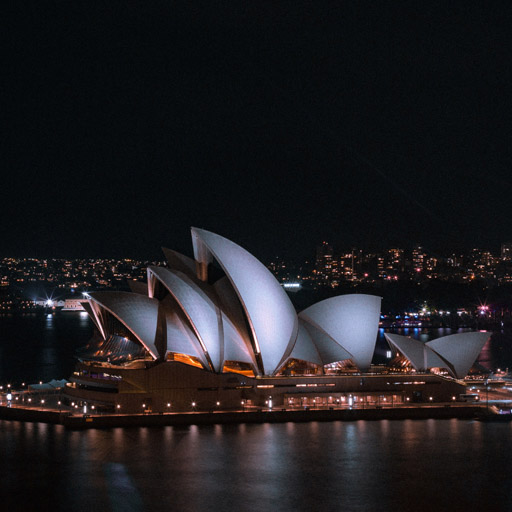}

\caption{Examples of photos in different time slots in a day: daytime, golden hour, blue hour, and nighttime. Photos by Unsplash users \texttt{lisanto\_12, bartmynameisbart, lanceanderson,  christopher\_\_burns}. }
\label{fig:data_example}
\end{figure}

\begin{figure}[t]
\centering
\includegraphics[width=0.32\linewidth]{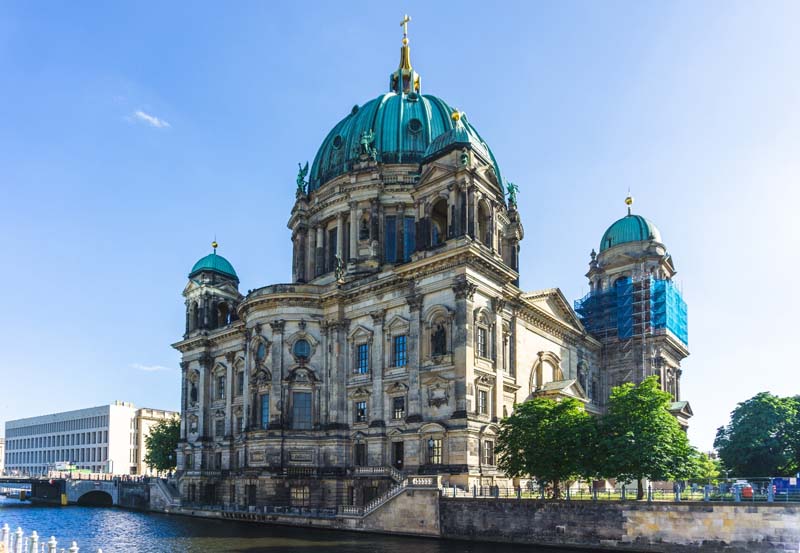}
\vspace{0.2em} 
\includegraphics[width=0.32\linewidth]{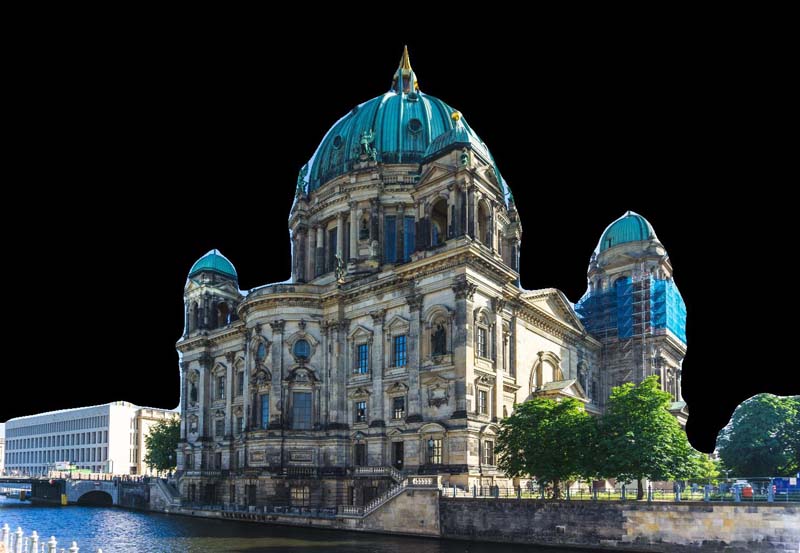}
\includegraphics[width=0.32\linewidth]{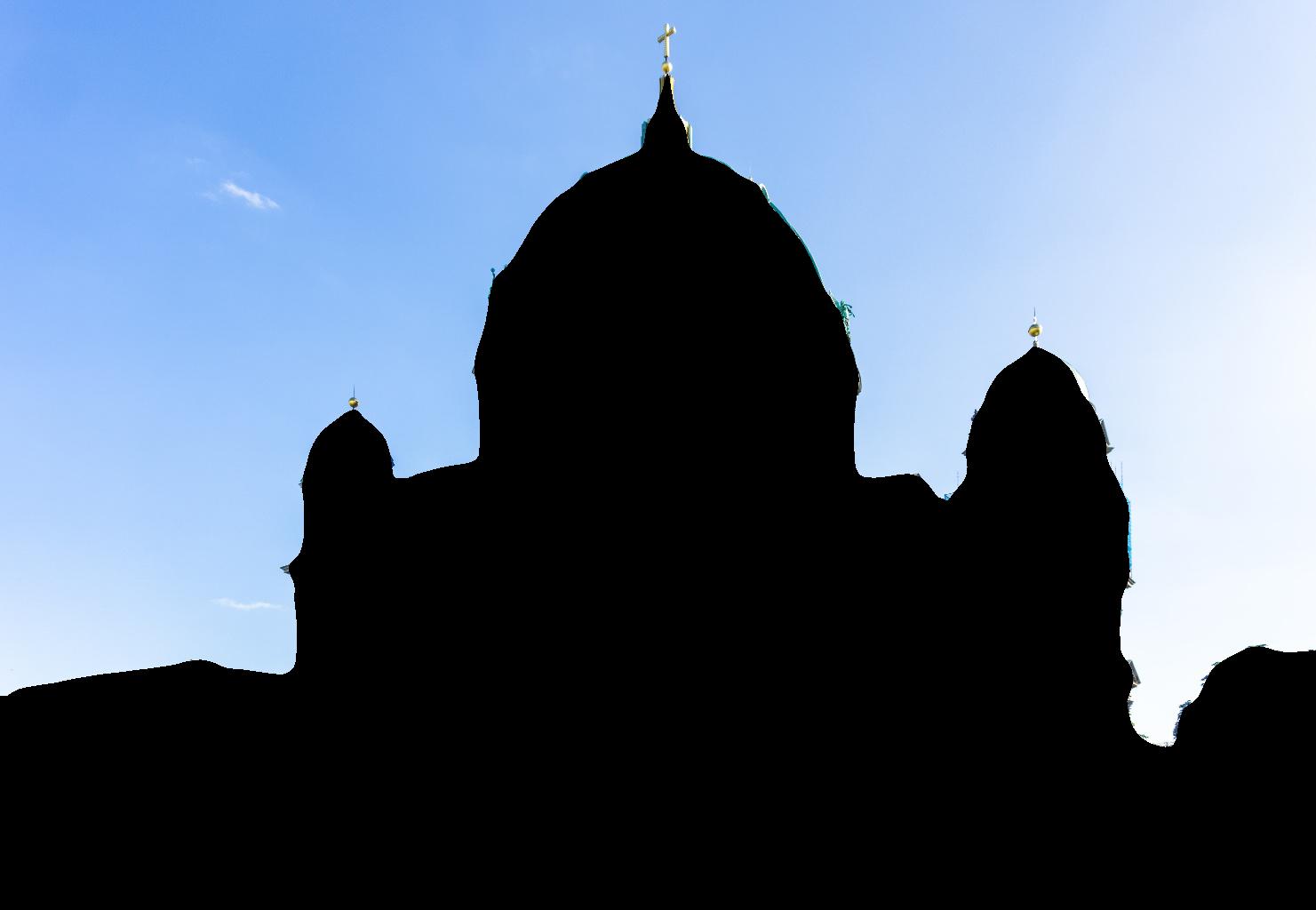}

\includegraphics[width=0.32\linewidth]{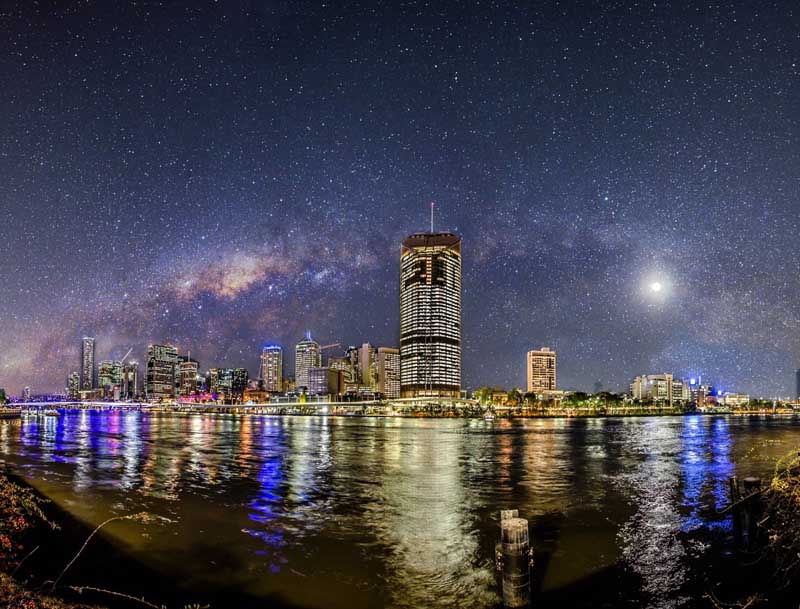}
\includegraphics[width=0.32\linewidth]{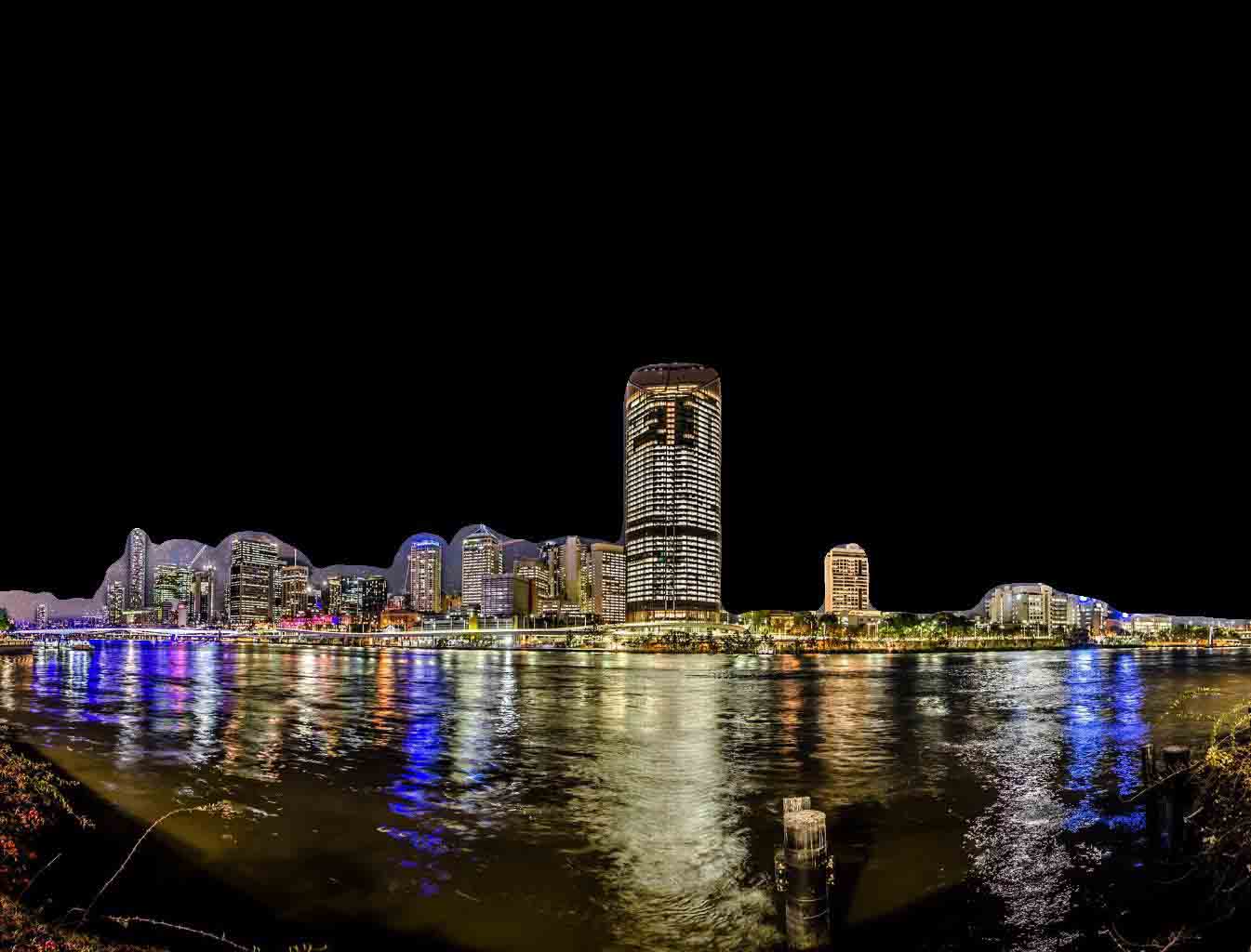}
\includegraphics[width=0.32\linewidth]{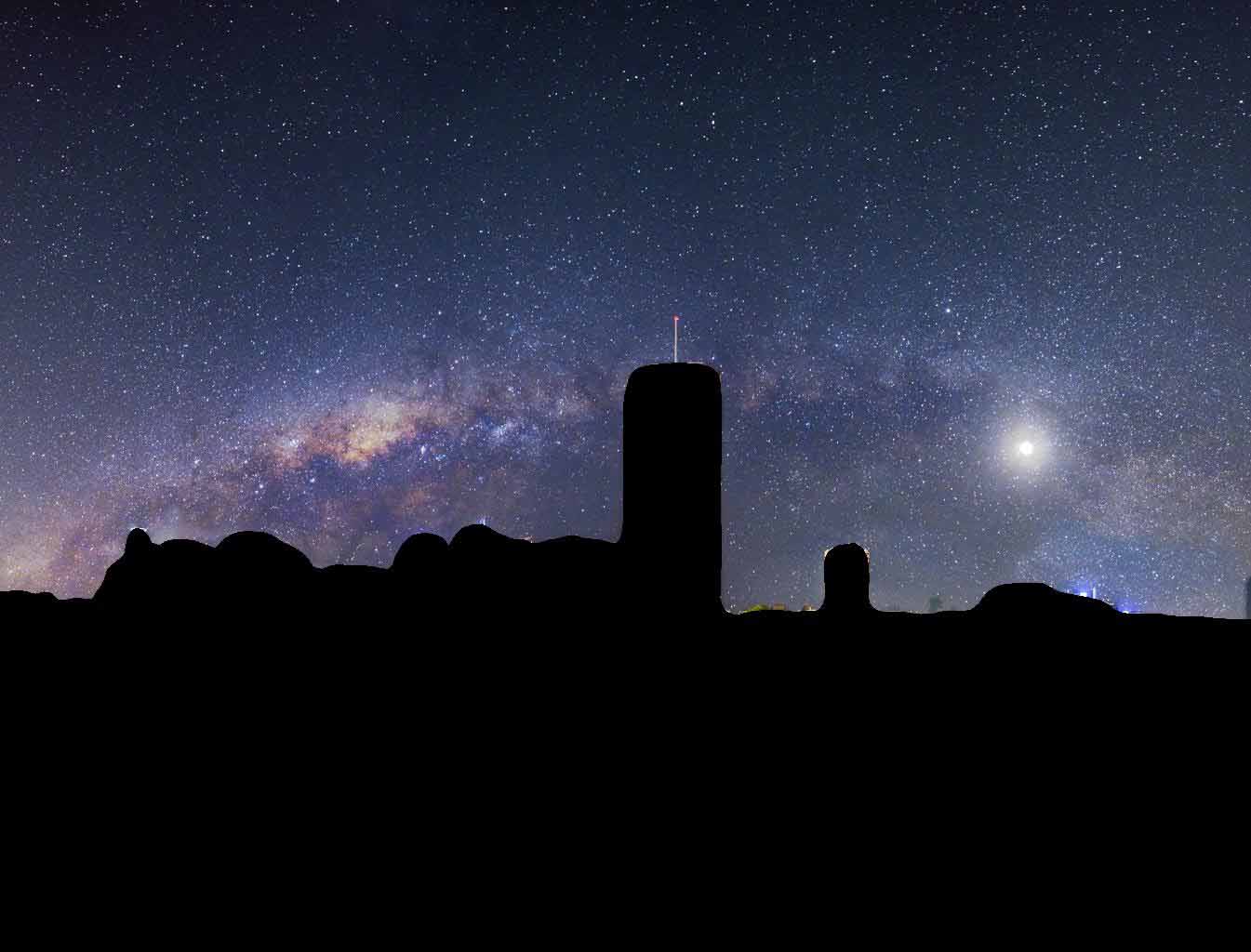}

\caption{Examples of segmentation of an outdoor architectural scene. Photos by Unsplash users \texttt{jose\_maria\_sava, michael75}}
\label{fig:data_seg}
\end{figure}

Some data and segmentation examples are shown in Figure \ref{fig:data_example} and \ref{fig:data_seg}. From day to night, illumination sources keep changing. In the daytime, the illumination of buildings comes from the ambient environment, mainly the sky.  When the sun is falling and the sky is becoming dark, buildings derive lighting sources from interior and exterior lights such as buildings and street lights. Daytime images can always provide sharp and detailed geometry of the scenes.  In contrast, images in golden, blue, and night hours depict pleasing and artistic joy in a day with mysterious chrominance and texture variation. For instance, at dusk and dawn (golden hours), the sky becomes yellowish-orange, and buildings are coated with warm daylight; at night, buildings are lightened with colorful glows or have inner gleaming lights on. 

\begin{table}[ht]
    \setlength{\tabcolsep}{5pt}
    \centering
    \begin{tabular}{l|rrrrr}
        \toprule
        source / label & \textbf{day} & \textbf{golden} & \textbf{blue} & \textbf{night} & \textbf{total} \\ 
        \midrule
        training set & 7,382 & 5,488 & 3,397 & 4,024 & 20,291\\
        
        \textit{unpaired imgs}   & 6,463 & 3,942 & \revised{2,567} & 2,936  & \revised{15,908}\\
        
        \textit{video frames} & 919  & 1,546  & 830 & 1,088  & 4,383\\
        
        \midrule
        evaluation set & 384 & 275 & 145 & 199 & 1,003 \\ 
        \bottomrule
    \end{tabular}
    
    \caption{Dahta statistics.}
    \label{tab:data_stat}
\end{table}

\section{Network Implementation Details} \label{sec:implement}
This section describes details about the network architecture and training settings in the image translation module.

\subsection{Network Architecture}
Our style transfer network consists of a content encoder $E^c$, a content specific domain mapping $M$, a style encoder $E^s$ and a generator $G$ for each domain. The generator can transfer style from source domain to target domain given source content and target style representations. For two domains $X_1$ and $X_2$, we train both transfer directions simultaneously, i.e., from $X_1$ to $X_2$ and from $X_2$ to $X_1$. 

The content encoder consists of three convolutional layers for down-sampling, four residual blocks, one shared residual block for both domains. A de-convolutional layer plus a convolutional layer is used for domain mapping. Style encoder is designed with five convolutional layers followed by a global average pooling and a fully-connected layer. The length of style latent code is 8. Our generator contains four residual blocks, and 5 layers of upsample and convolutional layers. Our multi-scale discriminator has four convolution layers and one fully-connected layer on each scale. By default, we set it on three scales. We use Leaky ReLU activation for discriminator, the shared residual block in the content encoder, and use ReLU for style encoder, content encoder( except for the shared residual block), and generator except for the last de-convolutional layer to which tangent activation function is applied. We apply Instance Normalization (IN) to the content encoder and all residual blocks, Adaptive Instance Normalization (AdaIN) \cite{huang2017arbitrary} to residual blocks in the generator, and Layer Normalization to convolutional layers in the generator. 

Both foreground and background models use the same network architecture.

We use similar annotation to \cite{huang2018multimodal}, c7s1-64 stands for 7$\times$7 convolutional block with 64 filters and stride 1, uk denotes a 2 nearest-neighbor upsampling layer followed by a 5$\times$5 convolutional block with
k iterations and stride 1, rb, dc, GAP, fc stands for residual block, de-convolutional layer, global average pooling layer, and fully-connected layer.

Generation architecture details:  
\begin{itemize}  
    \item Content encoder $E^c$: c7s1-64, c4s2-128, c4s2-128, rb3s1-128 $\times$ 4, rb3s1-128 (shared)
    \item Domain mapping $M$: dc3s2-128, c4s2-128
    \item Style encoder $E^s$: c7s1-64, c4s2-128, c4s2-256$\times$3, GAP, fc8
    \item Decoder $G$: rb3s1-128$\times$4, u128, c5s1-2, u64, c5s1-2,  c7s1-3
\end{itemize}

Discriminator $D$ architecture  details: c4s2-64, c4s2-128, c4s2-256, c4s2-512, fc1

\section{Perceptual Study Details} \label{sec:user_study}

Two surveys were conducted for perceptual study, in terms of image photorealism, and structure similarity and style consistency. Photorealism indicates how much the image looks real as a photo. Structure similarity indicates how much the scene in the generated image looks the same as the scene in the input image. Style consistency indicates how accurate the color styles are transferred semantic accordingly (i.e. static foreground style to foreground, dynamic background style to background).

There are 73 participants in our user study. The participants are general audience with ages between 18 and 30. 
Before the participants started to fill out the questionnaires, they are required to read the brief introduction on style transfer with examples.

\begin{figure}[ht]
\centering
\includegraphics[width=0.70\linewidth]{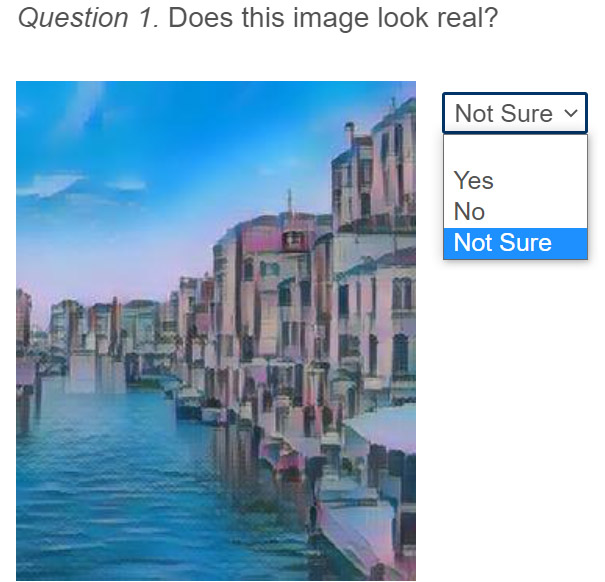}
\caption{Example of Photorealism Questions.}
\vspace{-1em} 
\label{fig:supp_user_study1}
\end{figure}

In the first questionnaire, we prepared 3 images in different target time slots (i.e., golden, blue hours, nighttime) generated from each baseline and our approach, and from the real world . In total there were (Ours) $\times$ 3 $+$ 10 (baselines) $\times$ 3 $+$ (real-world) $\times$ 3 $=$ 36 questions. As shown in Fig.~\ref{fig:supp_user_study1}, each question contains one image, and participants are asked if the displayed image is as realistic as a photograph. For each question, the participants need to select an option among \textit{"Yes"}, \textit{"No"} and \textit{"Not Sure"} for the same question:
\begin{quote}
    \textit{Q: "Does this image look real?"}    
\end{quote}

We accumulated the total number of each option and calculated percentages among all options for each method or the real-world group as the photorealism score. From result of photorealism scores in the main paper, our generated results look more realistic than other baselines.

\begin{figure}[ht]
\centering
\includegraphics[width=0.98\linewidth]{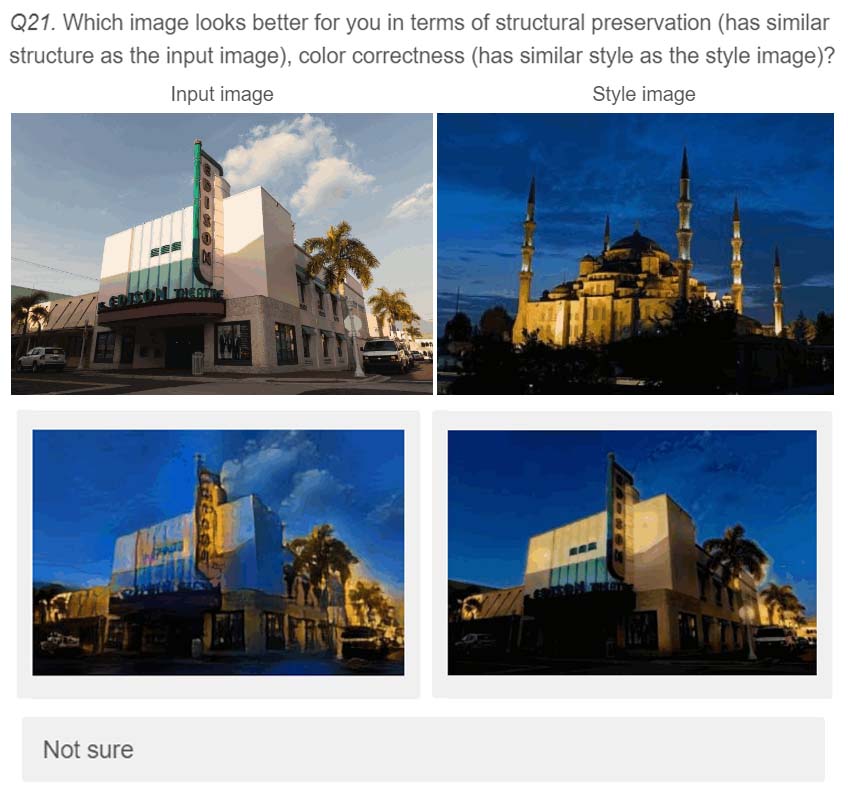}
\caption{Example of Pairwise Comparisons. Comparison between each baseline and our method in terms of structure similarity and style consistency.}
\vspace{-1em} 
\label{fig:supp_user_study2}
\end{figure}

In the second questionnaire, we conducted a perceptual study via pairwise comparisons between all baselines and ours in terms of structural preservation (i.e., generated images have similar structure as the input image) and color matching (i.e., generated images have similar style to the reference image). For each baseline, we prepared 6 comparison pairs comprised of 3 time slots by 2 different scenes. In total there were 10 (baselines) $ \times$ 6 (comparison pairs) $=$ 60 pair-wise questions in this questionaire. As shown in Fig.~\ref{fig:supp_user_study2}, each question contains an input image, a style image, and two results from a baseline and our method respectively. The participants are asked to select a better result (baseline's or ours) or an option of "\textit{Not sure}" for the question: 
\begin{quote}
    \textit{Q: "Which image looks better for you in terms of structural preservation (has similar structure as the input image), color correctness (has similar style as the style image)?"}
\end{quote}

As can be seen in pairwise results in main paper, our method outperforms the baselines, producing more natural matched color transfer and better structure preservation. 

After taking the surveys, participants were asked to give reasons and feedback about their choices.
We randomly picked unselected results in second survey and asked the participants to give reasons why they do not prefer these images.

The major comments  we received on deficient image quality in the user study are summarized as follows:
\begin{itemize} 
    \item Some edges in the images were broken or distorted.
    \item Some smooth image areas were destroyed or removed.
    \item There is wrong sunlight direction (e.g., the sun is at the backside of the building but the light reflection is at the front side of the building).
\end{itemize}

From these feedbacks, humans concern much edge or contour information and clear appearance for a photorealistic image. People can easily perceive stereo geometry from an image and illumination effect in real 3D world, which 2D-based image approach is hard to realize. Taking lighting direction into account for time-of-day style transfer can be an interesting future work. 

\begin{table*}[t!]
\renewcommand{\arraystretch}{1.3}
\setlength{\tabcolsep}{2.5pt}
\centering
\resizebox{0.98\textwidth}{!}{%

\begin{tabular}{c | c c c c c | c c c c | c c}
\toprule
 & DRIT++\cite{lee2018diverse,lee2020drit++} & MUNIT\cite{huang2018multimodal} & FUNIT\cite{liu2019few} & DSMAP\cite{chang2020domain} & StarGANv2\cite{choi2020stargan} & AdaIN\cite{huang2017arbitrary} & SANet\cite{park2019arbitrary} & AdaAttN\cite{liu2021adaattn} & LST\cite{li2019learning} & Ours & Ours-opt \\
\midrule
SSIM$\uparrow$ & 0.6093 & 0.5224 & 0.4373 & 0.4552 & 0.3645 & 0.4730 & 0.5150 & \underline{0.6538} & 0.4913 & 0.6371 & \textbf{0.7531} \\
e-SSIM$\uparrow$ & 0.5563 & 0.5061 & 0.4934 & 0.4779 & 0.4794 &  0.4979 & 0.4988 & 0.5411 & 0.4938 & \underline{0.6314} & \textbf{0.8200} \\
Acc$\uparrow$  & 92.88\% & 87.62\% & 83.30\% & \underline{94.61\%} & 73.22\% & 83.07\% & 92.78\% & 80.39\% & 89.69\% & \textbf{96.27\%} & 93.93\%  \\
IoU$\uparrow$ & 0.7182 & \underline{0.7533} & 0.5047 & 0.4936 & 0.3591 & 0.6676 & 0.7278 & 0.6543 & 0.6170  & 0.7362 & \textbf{0.7911} \\
\bottomrule
\end{tabular}
}
\caption{Evaluation results of Daytime to Golden Hour translation. \textbf{Bold} text indicates the best result; \underline{underlined} text indicates the 2nd best results.}
\label{table:exp_metrics_golden}
\end{table*}

\begin{table*}[t]
\renewcommand{\arraystretch}{1.3}
\setlength{\tabcolsep}{2.5pt}
\centering
\resizebox{0.98\textwidth}{!}{%

\begin{tabular}{c | c c c c c | c c c c | c c}
\toprule
 & DRIT++\cite{lee2018diverse,lee2020drit++} & MUNIT\cite{huang2018multimodal} & FUNIT\cite{liu2019few} & DSMAP\cite{chang2020domain} & StarGANv2\cite{choi2020stargan} & AdaIN\cite{huang2017arbitrary} & SANet\cite{park2019arbitrary} & AdaAttN\cite{liu2021adaattn} & LST\cite{li2019learning} & Ours & Ours-opt \\
\midrule
SSIM$\uparrow$ & 0.4211 & 0.3188 & 0.3905 & 0.4484 & 0.3349 & 0.4381 & 0.4894 & \underline{0.6324} & 0.4651 & 0.5735 & \textbf{0.6886}  \\
e-SSIM$\uparrow$ & 0.4881 & 0.5297 & 0.5010 & 0.5010 & 0.4767 &  0.5040 & 0.4924 & 0.5287 & 0.4974 & \underline{0.6309} & \textbf{0.8106} \\
Acc$\uparrow$  & 82.12\% & \underline{84.13\%} & 65.82\% & 82.96\% & 0.8378 & 64.05\% & 61.26\% & 46.80\% & 57.16\% & \textbf{91.13\%} & 83.81\%  \\
IoU$\uparrow$ & 0.7011 & 0.7236 & 0.5470 & 0.5344 & 0.3794 & 0.6797 & 0.7254 & 0.6733 & 0.6473 & \underline{0.7374} & \textbf{0.7915} \\
\bottomrule
\end{tabular}
}
\caption{Evaluation results of Daytime to Blue Hour translation. \textbf{Bold} text indicates the best result; \underline{underlined} text indicates the 2nd best results.}
\label{table:exp_metrics_blue}
\end{table*}

\begin{table*}[t]
\renewcommand{\arraystretch}{1.3}
\setlength{\tabcolsep}{2.5pt}
\centering
\resizebox{0.98\textwidth}{!}{%

\begin{tabular}{c | c c c c c | c c c c | c c}
\toprule
 & DRIT++\cite{lee2018diverse,lee2020drit++} & MUNIT\cite{huang2018multimodal} & FUNIT\cite{liu2019few} & DSMAP\cite{chang2020domain} & StarGANv2\cite{choi2020stargan} & AdaIN\cite{huang2017arbitrary} & SANet\cite{park2019arbitrary} & AdaAttN\cite{liu2021adaattn} & LST\cite{li2019learning} & Ours & Ours-opt \\
\midrule
SSIM$\uparrow$ & 0.0312 & 0.4598 & 0.2302 & 0.0919 & 0.1962 & 0.2761 & 0.4004 & \underline{0.5063} & 0.3251 & 0.4027 & \textbf{0.4806} \\
e-SSIM$\uparrow$ & 0.5198 & \underline{0.6600} & 0.4932 & 0.4581 & 0.4774 &  0.4866 & 0.4649 & 0.4883 & 0.4796 & 0.6453 & \textbf{0.7975} \\
Acc$\uparrow$  & 92.10\% & 88.60\% & 82.29\% & \underline{95.61\%} & 90.19\%  & 83.27\% & 41.45\% & 53.70\% & 74.57\% & \textbf{97.16\%} & 92.46\% \\
IoU$\uparrow$ & 0.6553 & \textbf{0.7378} & 0.5902 & 0.4647 & 0.4915 & 0.6452 & 0.7017 & 0.6321 & 0.6150 & 0.7034 & \underline{0.7318}\\
\bottomrule
\end{tabular}
}
\caption{Evaluation results of Daytime to Nighttime translation. \textbf{Bold} text indicates the best result; \underline{underlined} text indicates the 2nd best results.}
\label{table:exp_metrics_night}
\end{table*}

\begin{table}[!t]
\setlength{\tabcolsep}{1.1pt}
\centering
\resizebox{0.49\textwidth}{!}{
\begin{tabular}{l | c c c c | c }
\toprule
 & Acc-golden & Acc-blue & Acc-night  & Acc-mean & IS \\
\midrule
Eval  & 99.64\% & 98.62\% & 100\% & 99.42\% & 2.8340\\ 
\bottomrule
\end{tabular}
}  
\caption{Accuracy and IS of evaluation as reference}
\label{table:exp_eval_set}
\end{table}
\begin{table}[!t]
\setlength{\tabcolsep}{2.5pt}
\centering
\resizebox{0.49\textwidth}{!}{
\begin{tabular}{l | c  c  c  c  c }
\toprule
 & w/o $\mathcal{L}_{gd}+\mathcal{L}_{kl}$ & w/o $\mathcal{L}_{kl}$ & w/o $\mathcal{L}_{gd}$ & $\mathcal{L}_{total}$ &  $\mathcal{L}_{total}$(opt)\\
\midrule
golden & 0.4626 & 0.5502 & 0.5110 & \underline{0.6314} & \textbf{0.8200}\\
blue & 0.4753 & 0.5394 & 0.5201 & \underline{0.6309} & \textbf{0.8106}\\
night & 0.5020 & 0.5720 & 0.5165 & \underline{0.6453} & \textbf{0.7975}\\
mean & 0.4800 & 0.5539 & 0.5159 & \underline{0.6359} & \textbf{0.8094}\\
\bottomrule
\end{tabular}
}
\caption{Ablation study with edge-SSIM metric ($\uparrow$) on geometry losses.}
\label{table:exp_essim_metrics_geo_losses}
\end{table}

\section{Additional Results}\label{sec:result} 

\subsection{Metric Details}
We describe the implementation of our metrics in the following paragraphs. 
All metrics excluding IS are computed for three types of style transfers, i.e., daytime to golden hours, blue hours, and nighttime, and get average score. In the main manuscript, we only report the mean score of three style transfers. We supplement all results in Section \ref{sec:more_qual}.

\textbf{Translation accuracy}. InceptionV3 \cite{szegedy2016rethinking} 
classifier can classify images to different classes (domains).
Similar to \cite{liu2019few}, we trained an InceptionV3 classification model with our dataset for three classes (golden, blue and nighttime). Before training, we extended the data scale by random cropping to increase the accuracy. Accuracy rates of top-1 prediction by the trained classifier for each translation are computed. 
High accuracy indicates good style transfer to the target domain.

\textbf{Quality Diversity}. Inception score (IS) \cite{salimans2016improved} measures how realistic the generation is and how much variety of output in an objective computation way. The trained InceptionV3 classifier with our dataset is used to calculate the IS score for all generations (with three styles) by each method.

\textbf{Geometry Similarity}. To evaluate geometry similarity, the typical differentiable structural similarity (SSIM) index \cite{wang2004image} estimates the structural similarity in terms of luminance, contrast and structure between two images. Thus large luminance difference (e.g. day to night style transfer) downgrades SSIM measurement of geometry preservation. Instead of directly using SSIM, we calculate \textit{Edge Conditioned SSIM} (edge-SSIM). edge-SSIM computes image structural similarity (SSIM) between Canny edge detected maps \cite{canny1986computational} of images. It can well alleviates luminance influence on geometry (e.g., comparison in Tab.~\ref{table:exp_metrics_night}). To get edge-SSIM, we first obtain the edge map using Canny edge detection function in OpenCV, and then calculate the SSIM between input (daytime domain) and output (target domain) edge maps using the image processing toolbox scikit-image in Python.

\textbf{Semantic Segmentation Accuracy.}
For segmentation, IoU is the area of overlap divided by the area of union between the predicted segmentation and the ground truth. We compute IoU using the same pretrained segmentation model mentioned in the main manuscript between all generations and ground-truth daytime inputs. Then we get average score for each style transfer. High IoU means that generated images preserve good structure and recognizable real-world style for architectural photos.

\begin{figure*}[t]
\centering
\includegraphics[width=1\linewidth]{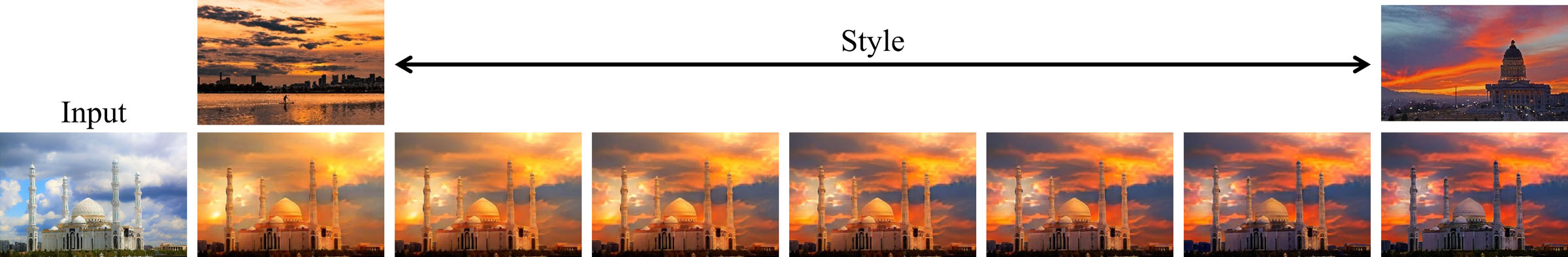}
\includegraphics[width=1\linewidth]{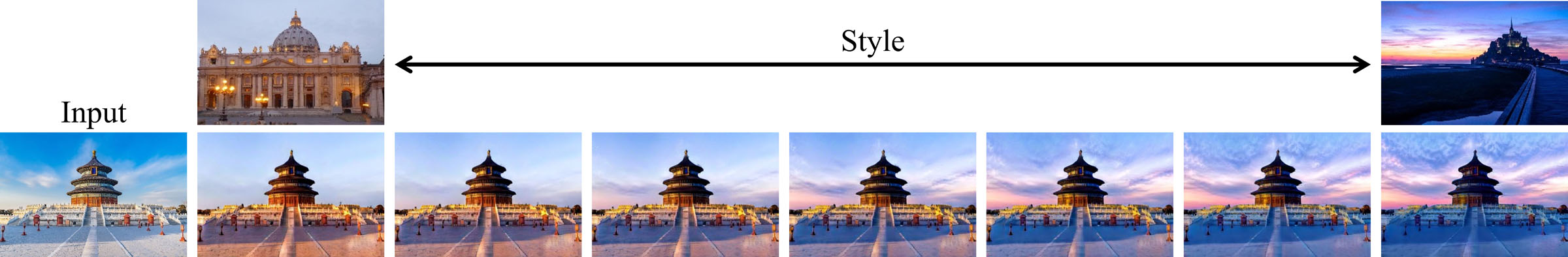}
\includegraphics[width=1\linewidth]{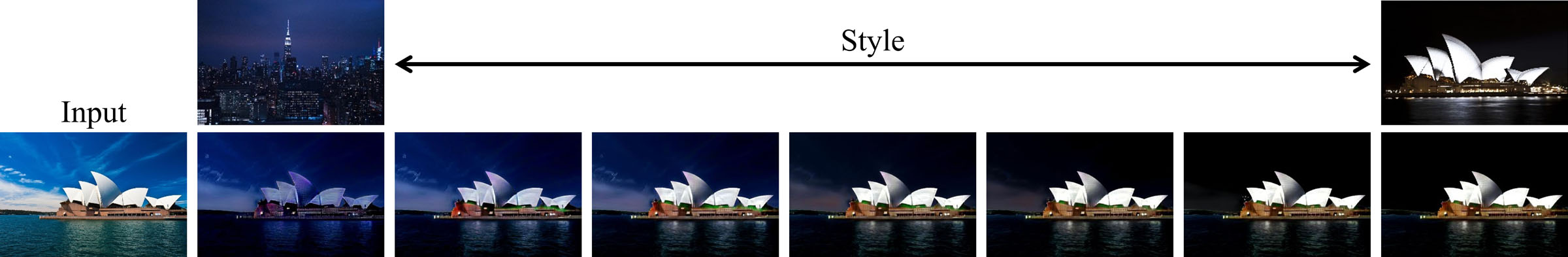}
\caption{Results of style interpolation.}
\vspace{-1em} 
\label{fig:qual_results_interpolation}
\end{figure*} 

\subsection{More Quantitative Results} \label{sec:more_qual}

\subsection*{Complete Quantitative Results}

We supplement complete quantitative results for all style translations, e.g., daytime to golden, blue and nighttime. The complete metric evaluation results among baselines and ours are displayed in Tables \ref{table:exp_metrics_golden}, \ref{table:exp_metrics_blue}, \ref{table:exp_metrics_night}, complete evaluation results of ablation study on geometry losses are shown in Table
\ref{table:exp_essim_metrics_geo_losses}. \textbf{Bold} texts are best results, \underline{underlined} texts are second best results. Table \ref{table:exp_eval_set} illustrates the accuracy and IS of the evaluation set for reference.

The traditional SSIM has unreasonable numbers in some cases such as day-to-night style transfer with dramatic luminance change (see SSIM in Fig.~\ref{table:exp_metrics_night}). Our edge-SSIM is more robust and stresses the edge information for structure similarity.

Overall, among image-to-image translation methods, MUNIT can retain primal input appearance (e.g., high edge-SSIM and IoU) but cannot transfer sufficient style in generated images (e.g., low accuracy, low IS). DSMAP accomplishes correct (high accuracy) and diverse style transfer (high IS),  but dramatically destroys original input geometry (low edge-SSIM, low IoU and unrecognizable visual results). StarGANv2 completely
destroys original appearance (please refer to the supplementary html viewer) with relatively low edge-SSIM and IoU. 

Neural style transfer approaches perform better on golden-style image generation (high classification accuracy for golden class), but fail to generate blue and night stylized images (low accuracy). Particularly, the state-of-the-art AdaAttN preserves more high-frequency geometry information (high edge-SSIM) than other neural style transfer methods but therefore somehow weakens its capability of accurate style transfer (low accuracy). AdaIN has better style transfer ability (high IS and high accuracy) but is bad at content preservation. Only our method can generate outputs with both high similarity of structure and appearance (high edge-SSIM), and semantically correct and sufficient style transfer (high IoU, accuracy, IS).

\subsection*{Running Time} 

Under training condition described in the main paper (i.e., same workstation), training time of \textit{MUNIT}, \textit{DRIT++}, \textit{DSMAP}, \textit{StarGANv2} or \textit{Ours} takes 2 to 3 days per model, and \textit{FUNIT}  takes over 4 days. For neural style transfer approaches, baseline models were trained for around 1 to 2 days.

To infer an image of 256$\times$, all baselines and ours take about 100ms to 200ms per image, and our image blending optimization takes around 300ms per image with 1 or 2 iterations. 
Some WCT$^2$ results are shown in the interactive viewer and we use pretrained WCT$^2$ model. By contrast, WCT$^2$ takes several seconds to predict an image of 256$\times$ or doubles the time for an image of 512$\times$.

\subsection{More Qualitative Results}

We show a complete visual comparison among different baselines. Please refer to the supplementary interactive html files to view the results.

\subsection*{Comparisons with I2I Translation Baselines}

In the interactive viewer, we show comparison results among DRIT++ (\cite{lee2018diverse,lee2020drit++}), MUNIT \cite{huang2018multimodal}, FUNIT \cite{liu2019few}, DSMAP \cite{chang2020domain}, and StarGANv2 \cite{choi2020stargan}.
FUNIT, DSMAP and StarGANv2 largely distort the building structure and appearance while  DRIT++, MUNIT can somehow preserve the geometry, but do not always have correct corresponding semantic style mapping. Ours keeps geometry information, also transfers sky style and texture, and transfers correct foreground color similar to reference style. Our blending optimized images recover much geometry detail with little  color loss. 

\subsection*{Comparisons with Neural Style Transfer Baselines}

State-of-the-art speedy neural style transfer methods (AdaIN~\cite{huang2017arbitrary}, SANet~\cite{park2019arbitrary}, LST~\cite{liu2020learning}, AdaAttN~\cite{liu2021adaattn}) tend to produce artistic effects with non-photorealistic texture and strokes even if we trained them with higher content (or other related) loss weight. Our results always tend to generate more photorealistic stylized images.

WCT$^2$ \cite{yoo2019photorealistic} takes much longer time at inference time (a few seconds) compared to other baselines and ours. Since it takes segmentation as input too, we use the same foreground and background mask for both WCT$^2$ and our method to achieve fair comparison. From results, WCT$^2$ perfectly preserves both foreground and background geometry but tends to transfer smooth color style for both segments. In general, the transferred style of WCT$^2$ is not as impressive as ours or baselines'.

\begin{figure}[t]
    \centering
    \small
    \def\sc{0.45}
    \begin{subfigure}[b]{\sc\linewidth}
        \centering
        \includegraphics[width=1\linewidth]{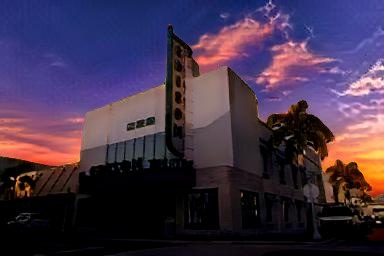}
        \vspace{-1.5\baselineskip} 
        \caption{Copy and Paste (Ours)}
    \end{subfigure}
    \begin{subfigure}[b]{\sc\linewidth}
        \centering
        \includegraphics[width=1\linewidth]{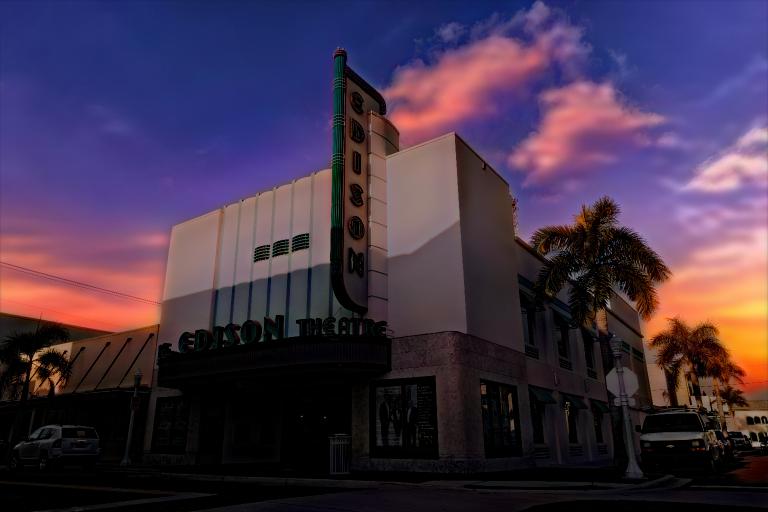}
        \vspace{-1.5\baselineskip} 
        \caption{Ours-opt}
    \end{subfigure}
    
    \vspace{0.1em}
    \begin{subfigure}[b]{\sc\linewidth}
        \centering
        \includegraphics[width=1\linewidth]{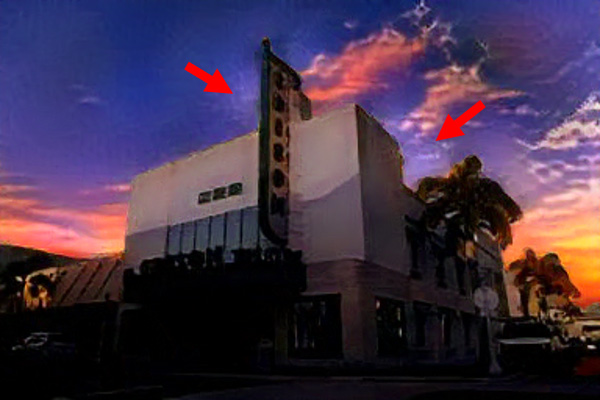}
        \vspace{-1.5\baselineskip} 
        \caption{DIB (bg)}
    \end{subfigure}
    \begin{subfigure}[b]{\sc\linewidth}
        \centering
        \includegraphics[width=1\linewidth]{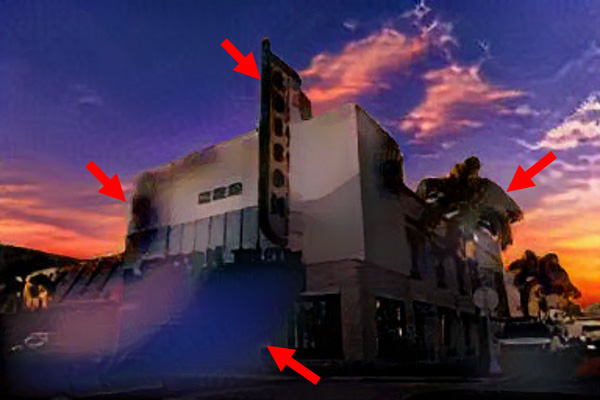}
        \vspace{-1.5\baselineskip} 
        \caption{DIB (fg)}
    \end{subfigure}
    
    \vspace{0.1em}
    \begin{subfigure}[b]{\sc\linewidth}
        \centering
        \includegraphics[width=1\linewidth,height=0.667\linewidth]{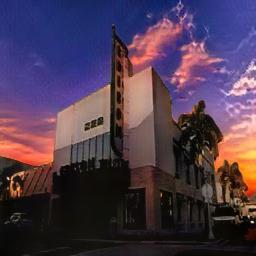}
        \vspace{-1.5\baselineskip} 
        \caption{DIH(fg)}
    \end{subfigure}
    \begin{subfigure}[b]{\sc\linewidth}
        \centering
        \includegraphics[width=1\linewidth,height=0.667\linewidth]{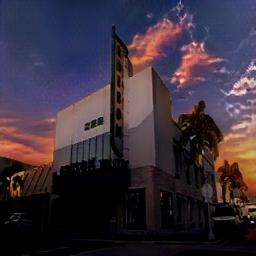}
        \vspace{-1.5\baselineskip} 
        \caption{DIH(bg)}
    \end{subfigure}
    \caption{Comparison to Blending and Harmonization Results. Deep Image Blending (DIB) uses (a) as target, (c) uses background image as source, (d) uses foreground as source. (e) and (f) are two Deep Image Harmonization (DIH) results. DIH is applied to \textit{Copy and Paste} composite images shown in top-right insets with according foreground and background segmentation masks. Best view with zoom.}
    \label{fig:blend_comp}
\end{figure}

\subsection*{Stylization Diversity}

Our models support style interpolation. We interpolate the outputs between two given style references by interpolating the style latent codes. The interpolation can generate smooth style transition as shown in the Fig.\ref{fig:qual_results_interpolation}.

\begin{figure*}[ht]
\centering
\includegraphics[width=1\linewidth]{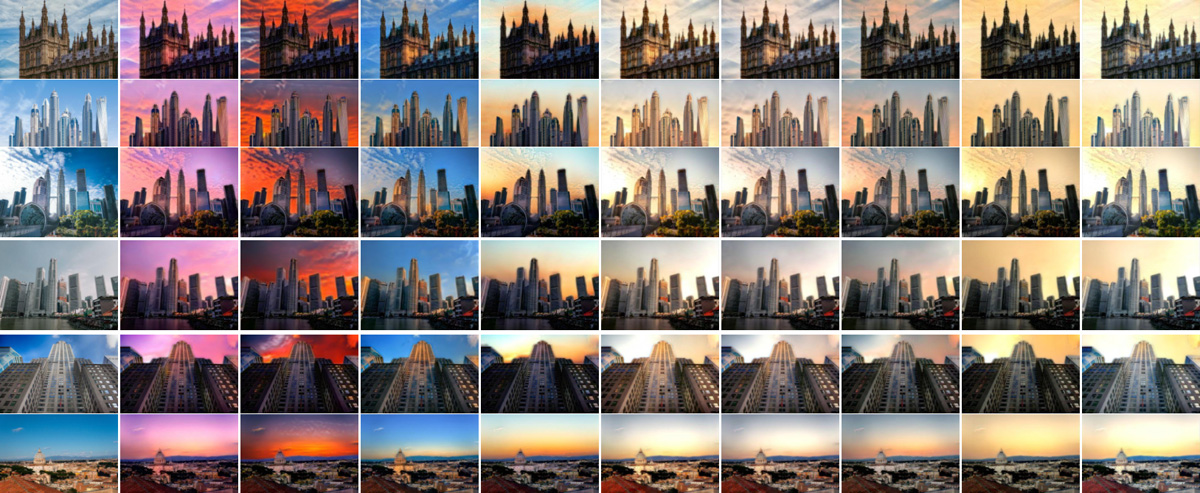}
\includegraphics[width=1\linewidth]{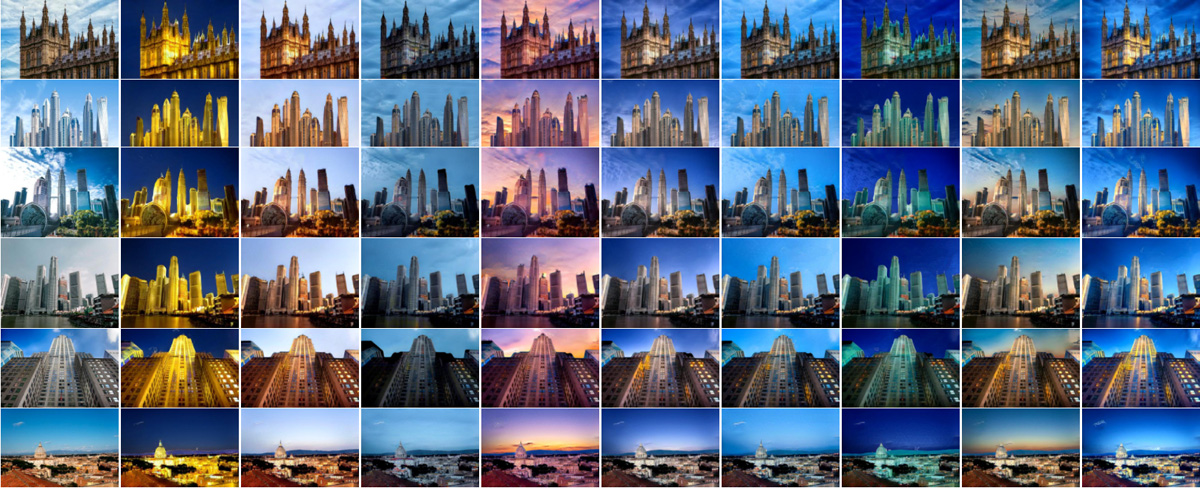}
\includegraphics[width=1\linewidth]{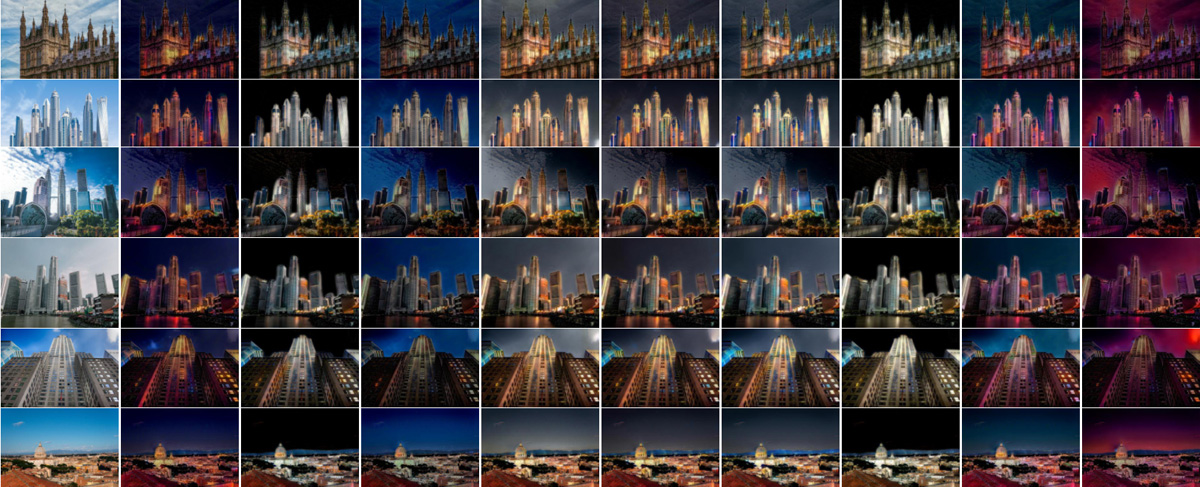}
\caption{Results of diverse styles under same scenes. Input in first column.}
\vspace{-1em} 
\label{fig:qual_results}
\end{figure*}

Fig.\ref{fig:qual_results} illustrates diverse style transfers given different style references. Our method support stylization across different architectural images and styles.
 
\subsection*{Comparisons of Blending and Harmonization Techniques}
To validate the effectiveness of our blending optimization, we show visual comparison of simple foreground and background addition \textit{Copy and Paste}, our blending optimized result, \textit{Deep Image Blending} (DIB) \cite{zhang2020deep} and \textit{Deep Image Harmonization} (DIH) \cite{tsai2017deep} results in Figure \ref{fig:blend_comp}. We used (a) \textit{Copy and Paste} result as target, foreground or background as source for DIB and DIH (results in (c-d) and (e-f)).
DIB tends to blend target style to source to make whole style consistent, which thus destroys original geometry or style. DIH changes building color or sky color according to the background and impair original transferred colors style (e.g., it brightens the building in (e) or darkens the sky in (f)). Our optimization approach (b) is able to refine geometry with original input gradient and preserve transferred colors.
Besides, DIB is slow and our method is about $5\times$ faster than DIB. On average DIB takes 2m14s to blend an image while ours takes 0.27s. 
Our blending optimization supports high resolution restoration unlike other blending or harmonization approaches. 

\end{document}